\documentclass[preprint,12pt,authoryear]{elsarticle}

\usepackage{amssymb}
\usepackage{inconsolata}

\usepackage{algorithmic}

\usepackage{multirow}
\usepackage{multicol}
\usepackage{times}
\usepackage{latexsym}
\usepackage{graphicx}
\usepackage{subfigure}
\usepackage{float}
\graphicspath{{images}}
\usepackage{stfloats}
\usepackage{booktabs}
\usepackage{indentfirst}
\usepackage{amsmath}
\usepackage{amssymb}

\usepackage{arydshln}

\usepackage{hyperref}
\usepackage{pifont}
\usepackage{color, colortbl}
\usepackage{comment}
\usepackage{subcaption}
\definecolor{Color}{gray}{0.9}

\usepackage{CJKutf8}

\usepackage{makecell}

\DeclareMathOperator*{\argmin}{\arg\!\min}

\usepackage[linesnumbered,ruled,vlined]{algorithm2e}

\SetKwInput{KwInput}{Input}             
\SetKwInput{KwOutput}{Output}           

\journal{AIIM}

\begin{document}

\begin{frontmatter}



\title{Text2MDT: Extracting Medical Decision Trees from Medical Texts}


\author[a]{Wei Zhu\footnote{Corresponding Author. Email: wzhu@stu.ecnu.edu.cn. Address: Department of Computer Science and Technology, East China Normal University, No. 3663, Zhongshan Road, Putuo District, Shanghai, China, 200050. }}
\author[a]{Wenfeng Li}
\author[c]{Xing Tian}
\author[a]{Pengfei Wang}
\author[a]{Xiaoling Wang}
\author[b]{Jin Chen}
\author[a]{Yuanbin Wu}
\author[c]{Yuan Ni}
\author[c]{Guotong Xie}

\affiliation[a]{organization={Department of Computer Science and Technology, East China Normal University},
            state={Shanghai},
            country={China}}

\affiliation[b]{organization={University of Kentucky},
            state={Kentucky},
            country={United States}}

\affiliation[c]{organization={Pingan Health Tecnology},
            state={Shanghai},
            country={China}}

\begin{abstract}

Knowledge of the medical decision process, which can be modeled as medical decision trees (MDTs), is critical to build clinical decision support systems. However, the current MDT construction methods rely heavily on time-consuming and laborious manual annotation. In this work, we propose a novel task, Text2MDT, to explore the automatic extraction of MDTs from medical texts such as medical guidelines and textbooks. We normalize the form of the MDT and create an annotated Text-to-MDT dataset in Chinese with the participation of medical experts. We investigate two different methods for the Text2MDT tasks: (a) an end-to-end framework which only relies on a GPT style large language models (LLM) instruction tuning to generate all the node information and tree structures. (b) The pipeline framework which decomposes the Text2MDT task to three subtasks. Experiments on our Text2MDT dataset demonstrate that: (a) the end-to-end method basd on LLMs (7B parameters or larger) show promising results, and successfully outperform the pipeline methods. (b) The chain-of-thought (COT) prompting method \cite{Wei2022ChainOT} can improve the performance of the fine-tuned LLMs on the Text2MDT test set. (c) the lightweight pipelined method based on encoder-based pretrained models can perform comparably with LLMs with model complexity two magnititudes smaller. Our Text2MDT dataset is open-sourced at \url{https://tianchi.aliyun.com/dataset/95414}, and the source codes are open-sourced at \url{https://github.com/michael-wzhu/text2dt}.

\end{abstract}







\end{frontmatter}


\begin{CJK*}{UTF8}{gbsn}


\section{Introduction}\label{sec:introduction}

As a typical application of artificial intelligence in the medical field, clinical decision support systems (CDSS) have been widely concerned by researchers \cite{DBLP:journals/isci/Tsumoto98,2006Clinical,MACHADO201721}. CDSS can suggest experienced doctors of all the options and problems to be considered when making decisions, help inexperienced medical students to learn clinical knowledge, or give medical advice to patients without medical background \cite{IoannisVourgidis2018Medical}. The core of building a CDSS is the knowledge of medical decision processes, which are rules that link given conditions to medical decisions \cite{2005Rule} and are usually modeled as medical decision trees (MDTs). However, existing methods for constructing MDTs rely on manual tree construction by medical experts \cite{SAIBENE2021114900}, which is time-consuming, laborious, and cannot absorb the latest research timely. All these hinder the construction, dissemination, maintenance  of large-scale CDSS \cite{key:article}. There is an unmet need to explore automated pipelines to precisely extract MDTs from vast and rapidly growing medical knowledge sources.

\begin{figure*}[ht]
\begin{center}
\includegraphics[width=0.96\textwidth]{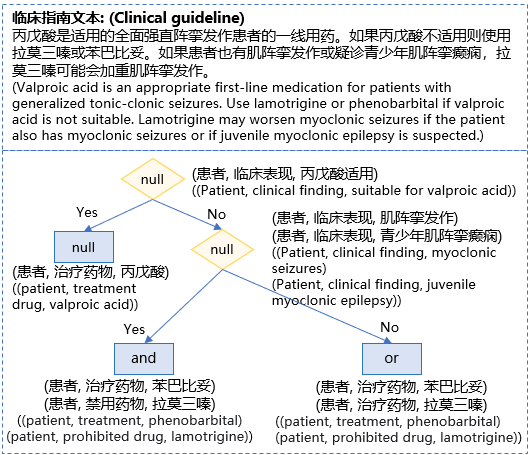}
\end{center}
\caption{An example of a medical decision tree contained in a medical text from an epilepsy clinical guideline. English translation are provided in brackets.   }
\label{fig:text2dt_example_1}
\end{figure*}

It is computationally challenging to automatically extract MDTs for the following reasons: 1) the current MDT lacks a normalized and structured form, leading to ambiguity in understanding medical decision knowledge and therefore hinders automated knowledge extraction; 2) the NLP community lacks a benchmark dataset for training and validating MDT extraction tasks; and constructing such data is challenging in that annotating medical decision trees requires in-depth domain knowledge; 3) existing methods for medical information extraction are not directly applicable for MDT extraction.

In this work, Text2MDT is defined as an automated task to explore the automatic extraction of MDTs from medical texts such as medical guidelines and textbooks. To this end, we structure and normalize a specific tree structure to model medical decision knowledge. As shown in Figure \ref{fig:text2dt_example_1}, the knowledge of a medical decision process embedded in the medical text can be modeled as a binary decision tree consisting of condition nodes (orange diamond) and decision nodes (blue rectangle). The triplets with their logic relationships in the nodes represent the conditional judgment to be performed or the decisions to be made based on the previous conditional judgment. Suppose the result of the conditional judgment is "Yes" ("No"), go to the left (right) branch for the following conditional judgment or decision. Once a decision is made, the medical process is terminated. In summary, the new MDT reflects the triplets in the text that represent medical knowledge and connects this information to form a complete decision-making process. We construct the first Text-to-MDT (Text2MDT) becnhmark dataset with 500 Text2MDT pairs and 3,232 triplets for automatically extracting MDTs from medical texts. Medical guidelines and textbooks, which are referential for clinical decision making, are used as knowledge sources. Trigger words and templates customized by medical experts are used to locate medical text fragments that contain knowledge for clinical decision making. Finally, well-trained annotators and medical experts complete the annotation manually. 

With the constructed Text2MDT benchmark, we conduct a systematic evaluation of different pretrained model based methods. The first cohort of methods we consider is from the pipeline framework, in which the Text2MDT task is decomposed into three subtasks: triplet extraction, node grouping and tree assembling. Note that the existing encoder-based information extraction research can not be directly applied to deal with our novel Text2MDT task in a end-to-end (end2end) fashion, they can be applied in each subtask. We also consider generation-based methods for the pipeline framework. The second cohort of methods are all end2end methods. For the end2end framework, we mainly consider utilizing the generation capabilities of the pretrained generative LMs, especially the current large language models (LLMs). Notably, the chain-of-thought \cite{Wei2022ChainOT} (COT) style reasoning is investigated, which demonstrates to be beneficial. Experiments on our Text2MDT benchmark show promising results.

In summary, the main contributions of this work are: 
\begin{itemize}
\item We propose a well-defined novel task Text2MDT aiming to automatically extract MDTs from medical text. 
\item We construct the first Text2MDT benchmark dataset with the help medical practitioners. 
\item Both the pipelined and end2end models are investigated, including encoder-based methods and LLM fine-tuning. The experiments show that LLMs can perform strongly on our Text2MDT benchmark, however the encoder-based models can also perform comparably by utilizing a series of models dealing with different subtasks in the pipeline. 
\item The Text2MDT dataset and source codes are openly available\footnote{\url{https://github.com/michael-wzhu/text2dt}}, to facilitate future research. 
\end{itemize}

\section{Related Work}

\subsection{Medical natural language processing}

The developments in neural networks and natural language processing has advanced the field of medical natural language processing (MedNLP) \cite{2021arXiv211015803Z,hahn2020medical,zhu-etal-2021-discovering}. In the pre-BERT era, firstly, RNNs like LSTM/GRU are used for processing sequential medical data such as text and speech \cite{beeksma2019predicting}. Convolutional networks are also used for medical text classificaiton \cite{hughes2017medical}. The techniques of Graph neural networks are also explored for diagnose recommendations \cite{li2020graph}. In this period, many different model architectures are specially designed for better performances on a specific MedNLP task \cite{zhu-etal-2021-discovering,autotrans,Zhang2021AutomaticSN}. Since BERT \cite{devlin2018bert}, the pretrained language models (PLMs) become the deafult solution for MedNLP. In this stage, researchers become less interested in modifying the model architecture, but instead trying to pretrain or further pretrain a PLM from the open domain to the medical domain \cite{guo2021global,zhu-2021-mvp,pubmedbert}. With the wide study of LLMs, the field of MedNLP is also being revolutionized. There are already works on adapting LLM backbones to the medical domain question answering \cite{zhu2023ChatMed}. And \cite{PromptCBLUE} propose PromptCBLUE, a prompt learning based benchmark dataset for examing the LLMs' ability in MedNLP tasks. This work can also serve as a testbed for the current commercial or open-sourced LLMs, since the complexity of our novel task will pose 
great challenges for them.

\subsection{Information extraction from medical texts}

Information Extraction (IE) is a research topic of long history that aims to extract structured knowledge or factual information from unstructured texts \cite{yang2022survey}. The field of IE includes a wide range of tasks, such as named entity recognition \cite{das-etal-2022-container,landolsi2023information}, relation extraction (RE) \cite{2020arXiv200910680Z,li2022sequence}, event extraction \cite{hsu2022degree}, aspect-level sentiment analysis \cite{cheng57aspect}. Since the raise of pre-trained models like BERT \cite{devlin2018bert}, the performances on IE tasks have advanced greatly \cite{zhu-2021-mvp}. But one has to have different model structures for different fine-grained IE tasks, for instance, the SOTA nested NER models \cite{Zhang2022DeBiasFG} are different from those of discontinuous NER tasks \cite{zhang-etal-2022-de}. Recently, there is a trend that all the IE task should be solved by a unified paradigm, that is, Seq2Seq generation. \cite{Yan2021AUG} proposes the framework of BartNER which solves all types of NER tasks with a BART model \cite{Lewis2019BARTDS}. UIE \cite{lu-etal-2022-unified} takes a step ahead and proposes to use prompts and a unified structural language to deal with many types of IE tasks with a single model checkpoint.

Medical information extraction is an important research field, and it has broad applications like medical search engine, automatic electronic health record analysis, online health consultation, and medical knowledge graph construction \cite{jmirMeidcalKnowledge,guo-etal-2021-global,zhu-etal-2019-panlp,Zhou2019AnalysisOT,zhu-etal-2021-discovering,Zhu2021pahtnlpM,Zhang2023NAGNERAU}. Compared with open-domain IE tasks, the IE tasks are known for their complexity. For example, discontinuous or nested entities are common in the medical field. And knowledge in the medical domain may be too complex to be expressed as triplets \cite{text2dt_shared_task}. For example, \cite{2019condition} introduced the role of “condition” and argued that a fact triplet is established based on some conditional triplets in the biomedical field. In the CMedCausal \cite{cmedcausal} task, a triplet may be the result of a subject conducting certain behaviour, expressing the causal relations. With the rise of LLMs, the research field of IE and medical IE is also under revolution. In this work, we compliment the existing literature by constructing the challenging Text2MDT task, where not only triplets have to be extracted, but also they need to arranged into nodes of a binary tree to express a complex medical decision process.




\subsection	{Text2Tree modeling}

There are a rich history of NLP tasks that aim to extract tree structures from a given text. The most fundamental task in NLP is syntax analysis, which aims to express the syntactic structure of a sentence into a syntactic tree \cite{Zhang2020ASO}. Parsing often relies on specific grammars, which are used to refine the output structures of syntax and semantics. Two of the most popular grammars are constituent parsing and dependency parsing. Text2Tree are also seen in many application scenarios. Math word problems (MWPs) \cite{Zhang2022MultiViewRC,Zhao2023AutomaticMS} extracts mathematical expressions from the unstructured texts, and try to improve the neural networks' capabilities in math problem solving by asking the model to understand the tree structure. Semantic parsing \cite{Kamath2018ASO}, the task of transforming the unstructured text into a SQL query, has promising application potentials in areas like dialogue systems, search engine, business intelligence. Our Text2MDT task is novel compared to the literature in the following sense: (a) Text2MDT focus on extracting medical decision trees from unstructured medical texts. (b) our task has a different granularity with the existing Text2Tree tasks, since each node in our task consists of one or more triplets. (c) the tree structure, or the links among different nodes, have different meanings with the existing Text2Tree tasks.

In terms of the model architectures for the existing Text2Tree methods, we have seen a trend of idiosyncratic models to more unified model architectures. The field of syntactic analysis has seen many different model architectures, such as recursive neural network \cite{Socher2011ParsingNS}, CRF \cite{Sutton2010AnIT}, transition-based models like \cite{fernandez-astudillo-etal-2020-transition,zhang-etal-2016-transition}, graph-based models \cite{Pei2015AnEN}. With the rise of pre-trained encoder models \cite{devlin-etal-2019-bert}, a series of works apply the pre-trained models like BERT to enhance the performances on the Text2Tree tasks. For example, \cite{2016Deep} proposes to install the biaffine module on top of a pre-trained BERT for the dependency parsing task. This method models the relations among token pairs as a table-filling task and decode the tree structures of the entire input sequence in one forward pass. With the advances of generative language models, many works apply the pretrained sequence-to-sequence (Seq2Seq) models or GPT style models to Text2Tree tasks \cite{seq-math,seq-sql}. Since the generative models generate sequences that ignore the constraints of the tree, a series of approaches \cite{math-tree,sql-tree} are devoted to add constraints for tree-structured decoders by utilizing the structural information or syntactic rules. In this work, we contribute to the existing literature by conducting a systematic evaluation of the encoder-based and generation-based methods, especially open-sourced generative LMs with different scales.

\section{Problem Definition}
\subsection	{Text2MDT Task}
As shown in Figure \ref{fig:text2dt_example_1}, the Text2MDT task focuses on extracting the MDT from a given text containing the medical decision process from medical guidelines or textbooks. We denote a medical text with $n\_text$ words as $X= [x_1,\,x_2,\,......,\,x_{n_{text}}]$,  the goal of Text2MDT is to generate the pre-order sequence of the nodes in the MDT $T= [N_1,\,N_2,\,......,\,N_{n_{node}}]$. The pre-order sequence of the nodes in the MDT can uniquely represent this tree, which we explain in detail in Section \ref{s3.2}.

\subsection	{Medical Decision Tree}
\label{s3.2}
\textbf{Node structure} \quad Nodes in a MDT consist of three parts: role, triplets, and logical relationship between triplets. We denote a node by
\begin{align}
\text{Node} & =\{\text{Role}, \, \text{Triplets}, \, \text{Logical\_Rel} \}, \nonumber\\
\text{Role} & = \Diamond \text{ or } \Box, \nonumber\\
\text{Triplets} & = (t_1,t_2, ..., t_{n_{tri}}), \nonumber\\
\text{Logical\_Rel} & = \text{AND}, \text{OR} \text{ or } \text{NULL}, \nonumber\\
\label{eq:node_structure}
\end{align}
where: (a) $\text{Role}$ denotes the role of the node. $\text{Role} = \Diamond$ means that the node is a condition node describing certain statuses of patients (presented as diamond-shaped nodes in Figure \ref{fig:text2dt_example_1}), while $\text{Role} = \Box$ means that the node is a decision node demonstrating how to treat the patients given certain conditions. (b) $\text{Triplets} = (t_1,t_2, ..., t_{n_{tri}})$ denotes the collection of triplets extracted from the given text, where each triplet $t=(sub,\,rel,\, obj)$ consists of a subject $sub$, a relation $rel$, and a object $obj$. These triplets are used to describe medical contents, either a patient' medical condition or status, or a medical decision representing the medical procedure to treat the patients. (c) $\text{Logical\_Rel}$ denotes the logical relationship (and/or relation) among the \text{Triplets} in a node. Note that $l=null$ if and only if the number of triplets $n_{tri}$ in the node is less or equal to 1.   
\textbf{Tree structure.} A medical decision tree represents the structured process for decision making of physicians. As depicted in Figure \ref{fig:text2dt_example_1}, medical professionals need to identify the condition of patients, and make the according decisions. Sometimes, medical conditions are complex so that one may have to differentiate many levels of conditions before they can make a valid medical decision. Therefore, we define a MDT as a binary tree consisting of condition and decision nodes, where non-leaf nodes are called conditional nodes, and leaf nodes are decision nodes. For the condition node, when the conditional judgment result is "Yes" ("No"), it will go to the left (right) branch for the next condition judgment or decision. It should be noted that each condition node has left and right child nodes. If the subsequent operation that needs to be done after the result of the condition judgment is "Yes" ("No") is not reflected in the text, a decision node without triplets is used as the left (right) child node. After this operation, a decision tree can be represented by a preorder sequence of its nodes. 

Figure \ref{fig:text2dt_example_1} shows a concrete example of MDT. In the example, the medical decision process embedded in the medical text above can be modeled by the MDT below: 1) Firstly, the condition "whether valproic acid is applicable for patients with generalized tonic-clonic seizures" is determined, and if the result is "Yes," i.e., valproic acid is applicable, then go to the left branch and make the corresponding decision, i.e., valproic acid is used for treatment; 2) if the result is "No," that is, valproic acid is not applicable, next go to the right branch and make another conditional judgment, i.e., the condition "whether the patient has myoclonic seizures or suspected juvenile myoclonic epilepsy" is determined, and go to different branches according to the result.

\section{Text2MDT Dataset}
\subsection{Data Collection}
We choose clinical practice guidelines and clinical medicine textbooks as our data sources. Clinical practice guidelines are systematically developed multidisciplinary clinical guidelines that help clinicians, patients, and other stakeholders make appropriate management, selection, and decisions about specific clinical issues. Clinical medicine textbooks are the primary means medical students acquire medical knowledge and can be used as a reference for clinical decision-making. We collected more than 100 clinical guidelines published by authoritative medical institutions about 30 clinical departments from 2011 to 2021 and undergraduate clinical medical textbooks published by People's Health Publishing House\footnote{\url{http://www.pph166.com/}. } to build our dataset.

Since medical texts are long and contain rich and various medical knowledge, we used section-based filtering and trigger/template-based filtering to locate segments of medical texts that have knowledge of medical decision process based on the analysis of medical texts and the help of specialized doctors. First, we selected the chapters with a high density of medical decision knowledge, such as "Treatment", "Drug Selection" and "Medical Solutions" in the source data. Then, we analyzed and summarized the structure and pattern of the medical decision text construct templates and trigger words for medical decision knowledge. We filtered the text based on the template and triggers to obtain the text fragments containing the knowledge of the medical decision process.

Annotators of our dataset include 4 annotators and 2 medical experts. All the annotators have linguistic knowledge and are instructed with detailed and formal annotation principles for at least two hours, including understanding the medical decision-making process, the judgment of logical relationships, and the annotation specifications of triplets and decision trees. Two well-trained annotators firstly independently annotated each text and revised the initial annotation after discussion. Medical experts will examine the revised annotation to avoid errors or omissions and make the ﬁnal decision. It is discarded if an annotation cannot be agreed upon in the discussion or is ambiguous. Furthermore, we calculate the Cohen’s Kappa \cite{Kappa} to measure the agreements between two annotators. The result of triplet annotation is 0.83, which indicates a high degree of consistency; the result of MDT annotation is 0.37, which indicates a degree of consistency.

\subsection{Data Statistics}

\begin{table}[tb!]
\centering
\resizebox{0.5\textwidth}{!}{
\begin{tabular}{ccc}
\hline

\textbf{Tree\_Depth}     & \textbf{Amount} & \textbf{Proportion}\\
\hline
2 & 134    & 26.80\%    \\
3  & 302   & 60.40\%    \\
4  & 64   & 12.80\%    \\\hline
\end{tabular}}
\caption{\label{tab:tree_depth_stats}Statistics of the medical decision tree in Text2MDT dataset.}
\end{table}

\begin{table}[tb!]
\centering
\resizebox{0.55\textwidth}{!}{
\begin{tabular}{ccc}
\hline

\textbf{Relation\_Name}     & \textbf{Amount} & \textbf{Proportion}\\
\hline
clinical\_feature & 1374    & 42.51\%    \\
therapeutic\_drug  & 910    & 28.15\%    \\
medical\_option  & 561    & 17.36\%    \\
usage\_or\_dosage & 222     & 6.87\%     \\
forbidden\_drug    & 83     & 2.57\%     \\
basic\_information & 82     & 2.54\%     \\
\hline
\end{tabular}}
\caption{\label{tab:relation_type_stats}Statistics of the triplet relations in Text2MDT dataset.}
\end{table}

Table \ref{tab:tree_depth_stats} reports the statistics of the tree depth in the Text2MDT dataset. There are 500 text-tree pairs in the Text2MDT dataset; the decision tree depth is 2 to 4, the average number of nodes per tree is 3.76, and the average number of triplets per tree is 6.46. There are 1896 nodes in the dataset, including 934 decision nodes, 962 conditional nodes, 476 “or” nodes, 367 “and” nodes, and 1053 “null” nodes.

Table \ref{tab:relation_type_stats} reports the statistics of the triplet relations in the Text2MDT dataset. Text2MDT dataset has 6 relationships, where the relationship "prohibited\_drug" only accounts for 2.57\% percent of the total number of triplets, so the dataset exhibits long-tailed distributions. It should be noted that the triplet extraction in Text2MDT has the problem of SingleEntityOverlap (SEO), i.e., triplets in medical text share a single entity. 

\subsection{Manual Evaluation of Medical Decision Trees}
To evaluate the quality of the annotated medical decision tree and whether it can help make medical decisions, we invited 10 medical practitioners and 10 people without a medical background to complete the following two evaluation tasks: 1) We observed the subjects' performance (accuracy and time spent) in answering medical decision problems of similar difficulty under different settings (with medical texts or decision trees as a reference). 2) We asked subjects to evaluate the ability of medical texts and decision trees to represent the medical decision process (completeness, readability, helpfulness).

Most of the subjects could answer the decision-making questions more accurately or faster with the help of the MDTs and thought that our annotated MDTs are more readable and helpful for understanding the knowledge of the medical decision process while providing a comprehensive representation of decision knowledge in medical texts. This demonstrates the quality of our annotation and the strength of the decision tree in terms of expressive power. Besides, we provide a detailed manual evaluation of MDT in the Appendix.

\begin{figure*}[ht]
\centering
\includegraphics[scale=0.55]{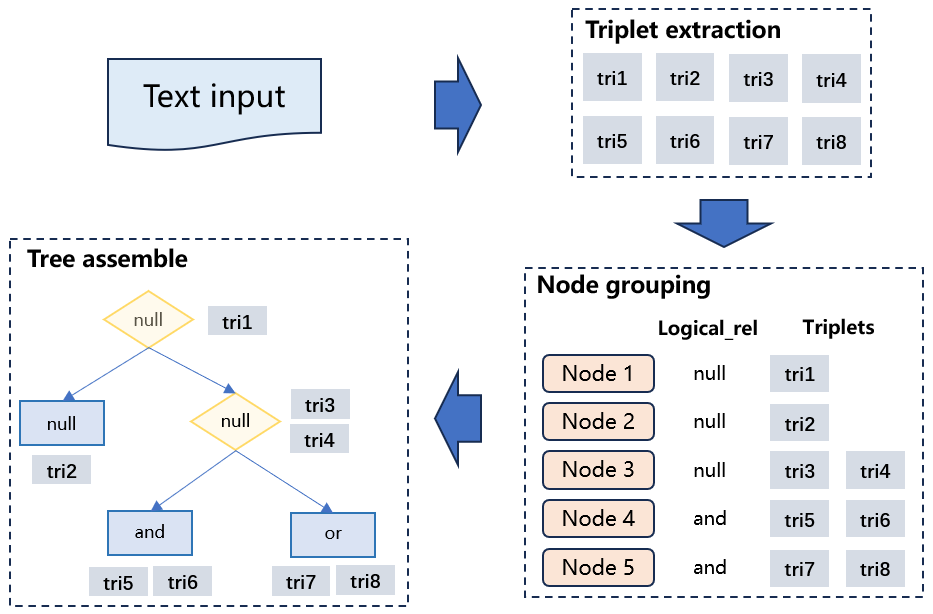}
\caption{Overview of our pipeline framework, which contains 3 subtasks: triplet extraction, node grouping and tree assembling.} 
\label{fig:pipeline_framework}
\end{figure*}

\section{Methods of modeling Text2MDT}
\label{sec:methods}

In this section, we will elaborate on our proposed methods for modeling the task of Text2MDT. First, we will present each module of the pipeline framework for Text2MDT. Then, we will discuss the end-to-end framework.  

\subsection{Pipelined framework}



Figure \ref{fig:pipeline_framework} demonstrates the pipeline for Text2MDT, which consists of three steps: triple extraction, node grouping, and tree assembling. 

\subsubsection{Triplet Extraction} \quad The first step is to extract all the triplets representing either decisions or conditions from medical texts with a unified triplet extraction model $\text{TEModel}()$: 
\begin{equation}
\{t_1, ..., t_{n_{tri}}\}=\text{TEModel}\left([x_1,\,......,\,x_{n_{text}}]\right),
\end{equation}
where $t_i=(s_i,r_i,o_i$) is the $i$-th triplet in the text, representing a part of a decision or a condition. $s_i$ and $o_i$ are two entity spans from the given text, and $r_i$ is a relation between the two entities and is one of the relation types presented in Table \ref{tab:relation_type_stats}.

Triplet extraction is widely studied task \cite{zhu-2021-autorc,Gao2023FPABEEFE,Zhu2021MVPBERTMP,autotrans}, and there are many recent works that can be utilized to complete this subtask. One line of work is based on semantic encoders like BERT \cite{devlin-etal-2019-bert} and a table-filling module \cite{dozat2016deep,Zhang2023FastNERSU}. The representative methods in this direction is: CASREL \cite{casrel}, TPLinker \cite{2020TPLinker} and UNIRE \cite{unire}. For completeness, we now demonstrate how UNIRE \cite{unire} applies a biaffine module to complete the entity mention detection and relation classification tasks simultaneously. 


With a given sentence input $X$, a pre-trained encoder like BERT or RoBERTa will encode the semantic information and provide hidden representations for $X^{'}$. Denote the hidden vector corresponding each token $x_i$ as $h_{i} \in \mathcal{R}^{d}$. Denote the set of entity types as $\mathcal{K}_{e}$, and the set of relation types as $\mathcal{K}_{r}$. UNIRE targets at identifying the label $l_{i, j}$ of each token pair $(i, j)$. That is, if the token pair $(i, j)$ is classified as an entity type $k_{e} \in \mathcal{K}_{e}$, we will consider the text span starting from the $i$-th token and ending at the $j$-th token as an entity of type $k_{e}$. And if the token pair $(i, j)$ is classified as an relation type $k_{r} \in \mathcal{K}_{r}$, and token $i$ and $j$ are the starting tokens of two entity mentions, we will consider that these two entities have a relation of type $k_{r}$. To complete the two tasks with a single calculation step, the UNIRE construct a biaffine module which maps each token pair $(i, j)$ to a probability distribution of dimension $K = |\mathcal{K}_{e}| + |\mathcal{K}_{r}| + 1$: \footnote{Adding 1 for the null type. } 
\begin{equation}
P(l_{i, j}) = \text{Biaffine}\left( h_i, h_j \right),
\label{eq:biaffine_module}
\end{equation}
where $\text{Biaffine}()$ is given by 
\begin{equation}
\text{Biaffine}(h_1, h_2)  =  h_1^{T} U h_2 + W \left( h_1 \oplus h_2 \right), 
\end{equation}
Since we need to calculate the scores for $K$ categories, $U$ is a $d \times K \times d$ tensor, and $W$ is a $2d \times K$ tensor.\footnote{Note that in the BERT biaffine NER \cite{yu-etal-2020-named}, two feed forward layers are designated to transform the two features passing to the biaffine module. However, we find that dropping the two feed forward layers will not result in significant performance changes. } Since the above method is analogeous as filling in a $n_{text} \times n_{text}$ sized table, we often refer to the biaffine method as the table-filling method. Denoting the ground truth of $l_{i, j}$ as $y_{i, j}$, then the training objective is the summation of cross-entropy loss at each of 
\begin{equation}
\mathcal{L}=-\frac{1}{|n_{text}|^{2}} \sum_{i=1}^{|n_{text}|} \sum_{j=1}^{|n_{text}|} \log P\left(l_{i, j} = y_{i, j}\right).
\end{equation}
After the above BERT-based biaffine model is trained, the inference procedure follows UNIRE \cite{unire}.



\subsubsection{Node grouping} \quad Given the medical text $X = [x_1,\,......,\,x_{n_{text}}]$ and the triplets $\{t_1, ..., t_{n_{tri}}\}$ extracted from this text, we now need to group these triplets into different groups, i.e., nodes, with relation $l \in (and, or, null)$ (a triple constitutes a group if it has the $null$ relation with other triples). These groups will be the main components of nodes of the MDT. 

Now we will demonstrate the model for this subtask: node-grouping biaffine (NG-Biaffine), which is to adapt the idea of biaffine model to the node grouping task. Note that if a triple belongs to a node with relation $l \in \mathcal{K}_{NG}$ (where $\mathcal{K}_{NG} = {\text{and}, \text{or}, \text{null}}$ is the set of the logical relations among triplets.), it will have relation $l$ with any other triplet within the group and $null$ relation with other triplets in the other groups. Thus, the key step for node grouping is to determine the relationships among the triplets, which can be conveniently modeled by a table-filling task similar to Equation \ref{eq:biaffine_module}. Denote the augmented text input as $X^{'} = [X, \text{[t]}, t_1, ..., \text{[t]}, t_{n_{tri}}]$, where $[]$ denotes the text concatenation operation. Note that we add a special token $\text{[t]}$ before each triplet. A pre-trained encoder like BERT or RoBERTa will encode the semantic information and provide hidden representations for $X^{'}$, and obtain the semantic representation of triplet $t_i$ by taking the hidden vector corresponding the special token right before $t_i$ (denoted as $h(t_i)$). Then a biaffine module will handle the classification task for each triplet pair $(t_i, t_j)$ by calculating its probability $P(l_{t_i, t_j})$ distribution over all the relation categories. 

During inference, we will consider a score based decoding procedure for resolving possible conflicts. For each triplet pair $(t_i, t_j)$, its label $l_{t_i, t_j}$ is obtained by choosing the relation category that receives the highest probability mass. And denote the probability mass of $l_{t_i, t_j}$ as $m_{t_i, t_j}$. During inference, we first calculate $m_{t_i, t_j}$ and $l_{t_i, t_j}$ for each triplet pair $(t_i, t_j)$ in a single forward pass. And we rank $l_{t_i, t_j}$ by $m_{t_i, t_j}$. The relation $l_{t_i, t_j}$ that receives the highest $m_{t_i, t_j}$ value will first be established, and any conflicting relation predictions with lower scores will be rejected. Here, a conflict arises when a triplet $t_i$ has the $and$ relation with $t_j$, but also has the $or$ relation with another triplet $t_{j^{'}}$. Then we will establish the relation prediction with the second highest probability mass that has not been discarded. Repeting the above procedures till all the triplets are included in the established relations, and we will have the complete prediction for node grouping. The logical relation for each node will be the relation type among the triplets inside the node.

\subsubsection{Tree assembling} \quad Note that in the above procedure, we already has the nodes in the decision tree. To assemble the nodes to a medical decision tree, one has to assign a role (condition or decision) to each node, and determine whether a pair of nodes are connected. Considering the node's role as the node's named entity label, and whether a pair of nodes are connected in the decision tree as a directional relation, the tree assembling task can also be regarded as a joint task of entity type classification and relation extraction. 

We now elaborate on the model details for tree assembling. Denote each unclassified node as $\text{Node}_{i}$ ($i$ = 1, 2, ..., $n_{node}$). We formulate each node as a text sequence by concatenating the logical relation name, role label name, and triplets' text contents, and we augment the text input $X$ to 
$$
X^{'} = [X, \text{[n]}, \text{Node}_{i}, ... \text{[n]}, \text{Node}_{n_{node}}]
$$, 
where $[]$ denotes the text concatenation operation. Note that we add a special token $\text{[n]}$ before each node. After being encoded with a pre-trained text encoder, we can obtain $h(\text{Node}_{i})$, the hidden states of the special token $\text{[n]}$ right before each node. $h(\text{Node}_{i})$ is considered as the semantic representation of $\text{Node}_{i}$. A simple linear layer can operate as the node type prediciton module, and a biaffine module will handle the relation classification task for each node pair $(\text{Node}_{i}, \text{Node}_{j})$. During decoding, we employ the strategy described in \cite{dozat2016deep} to resolve conflicting predictions. We will refer to the above model as TreeAssemble-Biaffine.

\subsection{LLM-based pipeline framework}

With the recent advances in generative language models \cite{PromptCBLUE,zhu-tan-2023-spt,Zhao2023ASO}, there is an unbreakable trend in the NLP field that transforms all the NLP tasks to the task of response generation given an prompt (or called instruction). \cite{yan-etal-2021-unified-generative} and \cite{Zhang2023NAGNERAU} propose to tackle the varous NER tasks to a unified generation framework. PromptCBLUE \cite{PromptCBLUE} conducts a thorough investigation on how the recent large language models perform on different medical text processing tasks, both under the in-context learning setting and instruction fine-tuning setting. 

Motivated by the above works, we now formulate each subtask of Text2MDT to a prompt-response generation task. In the Appendix, we present the prompt template and response format for each subtask in the pipeline framework. Note that for the generative LMs like LlaMA-2 to excel at each of the three tasks, we need to construct the designated datasets for each subtask, so that LMs can be finetuned. The details of constructing the instruction finetuning datasets are presented in Section \ref{sec:experiments}.



\subsection{End-to-end framework}

For the end2end framework, due to the complexity of this task, it is challenging for the encoder-based models to deal with the Text2MDT task in an end2end fashion. Thus, we mainly utilize the generative LMs for the end2end framework. Note that since this task is complex, it is natural that the idea of chain-of-thought (COT) \cite{Wei2022ChainOT} could be benefical for boosting the performances of the generative LMs. COT asks a generative LM to think step by step, either by itself (zero-shot COT), or to mimick the thinking steps of demonstrations (few-shot COT). In this task, we construction a series of different COT-style prompts and responses. 

Thus, for the end2end framework, we consider the following variations: 
\begin{itemize}
\item direct generation (Generation), in which a LM is asked to directly generate the entire MDT given the text inputs. The prompt and response templates are presented in the Appendix.
\item COT-style generation. Due to the complexity of our Text2MDT task, one can consider the following variations\footnote{See the Appendix for prompt and response templates for the series of generation methods. }:
\begin{itemize}
    \item COT-Generation-1, which decompose the Text2MDT task exactly as the pipeline framework, and ask the LM to first generate the extracted triplets, then node grouping, and then tree assembly, in a single generation run before generating the end-of-sentence token. 
    \item COT-Generation-2 decompose the task into more fine-grained subtask. It asks the model to generate entities, triplets, node assignments, node roles, and finally the entire tree. 
    \item COT-Generation-3 asks the LM to first extract triplets and then generate the whole MDT. 
    \item COT-Generation-4 decompose the triplet extraction subtask by asking the LM to first extract entities, and then generate the triplets, and finally generate the whole MDT. 
\end{itemize}
\end{itemize}

\section{Experiments}
\label{sec:experiments}
            
\subsection{Evaluation Metrics}














\subsubsection{Metrics for the triplet extraction subtask}

As described in Section \ref{sec:methods}, the most fundamental step of Text2MDT is to extract triples from the given text documents. Following \cite{PromptCBLUE} and \cite{zhu-2021-autorc}, we adopt the \textbf{triplet precision, recall and F1} scores as evaluation metrics. These metrics of triplet extraction are instance-level strict performance metrics. Here, an instance means a complete piece of information extracted from the given document. In our triplet extraction subtask, an instance consists of a head entity mention, a tail entity mention, and the relation label name between these two entities. And strict means that the model predicts an instance correctly if and only if it correctly predicts the all the components of the instance.

\subsubsection{Metrics for the node grouping subtask}

Following \cite{wang-cer-2012-stanford}, we now define an edit distance based metric to evaluate how models perform in the node assignment task. According to Equation \ref{eq:node_structure}, one can express a predicted node $N^{pred}$ to a tuple.
\begin{equation}
N^{pred} = (\text{Role}^{pred}, \, t_{1}^{pred}, \, ... , \, t_{n_{tri}}^{pred},  \, \text{Logical\_Rel}^{pred}).
\end{equation}
Note that we treat each triplet in the same level with the node role label and the logical relation label. And denote a node in the ground truth as 
\begin{equation}
N^{gt} = (\text{Role}^{gt}, \, t_{1}^{gt}, \, ... , \, t_{n_{tri}}^{gt}, \, \text{Logical\_Rel}^{gt}).
\end{equation}
Treating each element in the $N^{pred}$ and $N^{gt}$ tuples as indivisible, one can calculate the edit distance between $N^{pred}$ and $N^{gt}$. In this scenario, the editing operations include inserting and deleting elements, and each operation has a cost of 1. Now we concatenate all the nodes in the node grouping prediction into a single tuple $\text{NG\_Tup}^{pred}$. Since we does not require the model to assign orders to each node in the node grouping step, we consider all the permutation $m$ of nodes in the ground truth $\text{MDT}^{gt}$, and we concatenate the nodes in each permutation (denoted as $\text{NG\_Tup}^{gt, m}$). And the edit distance between the whole node assignment prediction and the ground truth node assignment is defined as the minimum edit distance between the predicted node grouping and a permutation of the ground truth node grouping:
\begin{align}
& \text{NG\_ED}(\text{NG\_Tup}^{pred}, \text{MDT}^{gt})   \nonumber \\
=  &  \min_{m \in \text{Permute}(\text{MDT}^{gt})} \text{ED}(\text{NG\_Tup}^{pred}, \text{NG\_Tup}^{gt, m}), 
\end{align}
where $\text{ED}(x, y)$ denotes the edit distance between tuple $x$ and tuple $y$. Since the edit distance score $\text{NG\_ED}$ is an un-normalized metric, it is in-suitable for model comparisons. Thus, we now define the Levenshtein ratio \cite{Navarro2001AGT} (denoted as NG\_LR) for the node grouping subtask: 
\begin{align}
&  \text{NG\_LR}(\text{NG\_Tup}^{pred}, \text{MDT}^{gt})  \nonumber \\
=  &  \dfrac{  \text{NG\_ED}(\text{NG\_Tup}^{pred}, \text{MDT}^{gt})  }{ \max ( \text{len}(\text{NG\_Tup}^{pred}),  \text{len}( \text{NG\_Tup}^{gt, m^{*}} ) }
\end{align}
where $\text{len}$ denotes the tuple length, and $m^{*}$ is the $\text{MDT}^{gt}$'s permutation that obtains the lowest edit distance with the prediction:
\begin{equation}
m^{*}  = \argmin_{m \in \text{Permute}(\text{MDT}^{gt}) } \text{ED}(\text{NG\_Tup}^{pred}, \text{NG\_Tup}^{gt, m}).
\end{equation}

\subsubsection{Metrics for the tree assembling subtask}

To properly evaluate a model's performance in constructing medical decision trees from text, we adopt the following three evaluation metrics: 
\begin{itemize}
\item The accuracy of decision tree extraction (TreeAcc). \quad For this metric, the instance is the entire medical decision tree consisting of a series of nodes connected as a binary tree of a certain structure, and each node contains three components, logical relation, role and triplets. A decision tree predicted by a model is correct when it is precisely the same as the ground truth. Thus, this metric is a very strict metric. 
\item F1 score of decision paths (DP-F1). \quad We define a decision path in a medical decision tree as a path from the root node to a leaf node. Thus, in DPF1, an instance is a decision path, and a model correctly predicts a decision path if and only if it correctly predicts all the nodes in the path and how they are connected. 
\item Lenvenshtein ratio of the decision tree (Tree\_LR). \quad Similar to the definition of edit ratio defined for the node grouping task, we can arrange the contents of all nodes in the predicted or ground-truth tree into a single tuple in the in the order of depth-first search (denoted as $\text{Tree\_Tup}^{pred}$ and $\text{Tree\_Tup}^{gt}$, respectively), and treat each triple, node role label, node logical relation as indivisible elements. Thus Tree\_LR is defined by 
\begin{align}
&  \text{Tree\_LR}(\text{Tree\_Tup}^{pred}, \text{Tree\_Tup}^{gt}) \nonumber \\
=  &  \dfrac{  \text{ED}(\text{Tree\_Tup}^{pred}, \text{Tree\_Tup}^{gt})  }{ \max ( \text{len}(\text{Tree\_Tup}^{pred}), \text{len}(\text{Tree\_Tup}^{gt})) }.
\end{align}

\end{itemize}

\subsection{Implementation Details}
\label{subsec:implement_details}

Our code was implemented with Pytorch\footnote{\url{https://pytorch.org/}.} and Huggingface Transformers\footnote{\url{https://github.com/huggingface/transformers}.}. For pretrained encoder based methods, we use the pre-trained Chinese medical BERT (denoted as MedBERT) by \cite{guo-etal-2021-global} as the default backbone model. For ablation studies, we also consider the widely used BERT-wwm-ext\footnote{\url{https://huggingface.co/hfl/chinese-bert-wwm-ext}.}, Google BERT-base Chinese \cite{devlin-etal-2019-bert}, and Erlangshen-ZEN1-224M-Chinese\footnote{\url{https://huggingface.co/IDEA-CCNL/Erlangshen-ZEN1-224M-Chinese}.}. For the decoding module such as the biaffine module \cite{dozat2016deep} and \cite{unire}, we will use the original authors' default configurations. We will fine-tune all the model parameters. Batch size is set to 8, warm-up steps is set to 50, the number of training epochs is set to 50, the learning rate is set to 2e-5 with a linear schedule, and the optimizer is AdamW \cite{Loshchilov2017FixingWD}. The other hyper-parameters like gradient clipping, Adam epsilon are kept the same with the Transformers repository. 


For generative LMs, we consider a collection of well-known language models of different sizes. (a) GPT-2 Chinese\footnote{https://huggingface.co/uer/gpt2-chinese-cluecorpussmall}. (b) Randeng-T5-784M\footnote{https://huggingface.co/IDEA-CCNL/Randeng-T5-784M-MultiTask-Chinese}. (c) BLOOMZ-7.1B-mt\footnote{https://huggingface.co/bigscience/bloomz-7b1-mt}. (d) ChatGLM-6B-2. (e) ChatMed\footnote{https://github.com/michael-wzhu/ChatMed}, which is adapted from the LlaMA-7B backbone. (f) Chinese-LLaMA-2 7B/13B\footnote{https://github.com/michael-wzhu/Chinese-LlaMA2}, which are the Chinese version of LlaMA-2 models \cite{Touvron2023Llama2O} from Meta. (g) Ziya-13B-medical\footnote{https://huggingface.co/shibing624/ziya-llama-13b-medical-lora}, which is also further pre-trained with the LlaMA-2 models. (h) Baichuan-2 7B/13B models\cite{Yang2023Baichuan2O}, which are one of the most recent open-sourced Chinese LLMs, and have achieved excellent performances in many evaluation benchmarks like \cite{li2023cmmlu}. Unless stated otherwise, We will use Baichuan-2 7B as the default generative LM. For generative LMs with parameters fewer than 500 millions, we fine-tune all the model parameters. For larger models, we will use LoRA \cite{hu2021lora} with rank 24 to modify the query, key, value, and output matrix in the self-attention module, and the two matrices in the feed-forward module \cite{PromptCBLUE}. The LoRA parameters are fine-tuned with learning rate 1e-4. The rest of the hyper-parameters are set the same with the encoder-based methods.  

For each method, we validate the model performance on the dev set and choose the checkpoint with the best dev performance to predict on the test set. Each experiment is run for 5 times with different random seeds and the average scores are reported.


\subsection{Datasets}

The annotated Text2MDT dataset is open-sourced at \url{https://github.com/michael-wzhu/text2dt}. It is one of the evaluation dataset for The CHIP-2022 shared tasks \cite{text2dt_shared_task}. And it is now a part of the CBLUE benchmark \cite{zhang-etal-2022-cblue} and PromptCBLUE \cite{PromptCBLUE} benchmark. The original Text2MDT has a 800:100:100 train/dev/test split. Since we are experimenting with different methods from the pipeline and end2end frameworks, we now need to construct different variations of the Text2MDT datasets. 

\subsubsection{Datasets for the pipeline framework}

Since the pipeline framework has three subtasks, thus, we need to construct a different dataset for each subtask so that we can train an encoder-based model: 
\begin{itemize}
\item Text2MDT-TE, the Text2MDT triplet extraction dataset, where the input is the medical text, and the target is the list of triplets in the structured format like JSON. This dataset has a 800:100:100 train/dev/test split.
\item Text2MDT-NG, the Text2MDT node grouping dataset, where the input is the medical text and the list of triplets in text sequence concatenated together, and the output is the list of nodes in the structured format like JSON and each node contains a list of triplets and a logical relation label. For the Text2MDT-NG training set, we augment the original Text2MDT four times by shuffling the orders of triplets. Thus, this dataset has a 3200:100:100 train/dev/test split.
\item Text2MDT-TA, the Text2MDT tree assembling dataset, where the input is the medical text and the list nodes in text sequence concatenated together, and the output is the list of MDT nodes in the structured format like JSON and each node contains a list of triplets, a logical relation label and a role label. For the Text2MDT-TA training set, we augment the original Text2MDT four times by shuffling the orders of nodes in the input. Thus, this dataset has a 3200:100:100 train/dev/test split.
\end{itemize}

For each of the above datasets, we will construct a prompt-based dataset for the generative LM methods, with the prompt and response templates in the the Appendix.

\subsubsection{Datasets for the end2end framework}

For each end2end method, we will construct the end2end dataset with the prompt and response templates in the the Appendix. So that each end2end dataset has a 800:100:100 train/dev/test split.

\subsection{Competing Methods}

\subsubsection{Methods for the pipelined framework} 

For the triplet extraction tasks, we consider the following methods: (a) UNIRE \cite{unire}; (b) TPLinker \cite{2020TPLinker}; (c) CasRel \cite{casrel}; (d) Sep-Biaffine, which divide the triplet extraction sub-task into two tasks, entity extraction and relation classification, and use two BERT-biaffine models to handle each task; (e) direct generation (Generation); (f) generation in the COT fashion (COT-Generation), which is to ask the LMs to first detects the relations in the given medical text and then extract triplets. The prompts for COT-Generation is presented in the Appendix. The first four methods use encoder-based PLMs and the last two utilize the generation capabilities of generative LMs. 

For the node grouping subtask, as discussed in Section \ref{sec:methods}, we could use the following methods: (a) the NG-biaffine method. (b) NG-TableFilling method, which substitute the biaffine module (Equation \ref{eq:biaffine_module}) in the NG-biaffine method to the table-filling module in \cite{2020TPLinker} (Equation (1) of \cite{2020TPLinker}). (c) generation based method (Generation), which asks a generative LM to directly generate the node grouping results given the text input. (d) COT style generation (COT-Generation), which asks the generative LM to first cluster the triplets into groups, and then identify the logical relation of each node. 

For the tree assembling subtask, we will use: (a) TreeAssemble-Biaffine method described in Section \ref{sec:methods}. (b) substituting the biaffine module of TreeAssemb to the the table-filling module of \cite{2020TPLinker} (Equation (1) of \cite{2020TPLinker}) results in the TreeAssemble-TableFilling method. (c) generation based method (Generation), which asks a generative LM to directly generate the medical decision tree given the text input. (d) COT style generation (COT-Generation), which asks the generative LM to first determine the role of each node, and then assemble the nodes to form a medical decision tree.


\subsubsection{Methods for the end2end framework} Following Section \ref{sec:methods}, we consider the following end2end methods: (a) Generation; (b) four variations of COT-style generation, (b1) COT-Generation-1; (b2) COT-Generation-2; (b3) COT-Generation-3; (b4) COT-Generation-4.

\subsection{Main experimental results}

In this subsection, we will report the experimental results for each subtask of the pipeline framework, and the overall performance when making predictions with the whole pipeline. In this part of the experiments, for encoder-based methods, we utilize the MedBERT from \cite{guo-etal-2021-global} as the pre-trained backbone. For generation-based methods, we utilize the Baichuan-2 7B as the backbone model.

\subsubsection{Performances on each subtask} \quad The results of each subtask are reported in Table \ref{tab:pipeline_framework_results}. From the results, we can see that: (1) Although two-magnitude smaller than the Baichuan-2 7B model in both parameters and complexity, the MedBERT (105 million parameters) based methods achieve competitive performances in all the subtasks, and only falls behind the COT-Generation method by 0.5\% in F1 on the triplet extraction task, 0.6\% in NG\_LR on the node group task, and 1.0\% in Tree\_LR on the tree assembling task. The disadvantages of encoder-based methods are that they require different decoding architectures for obtaining the final predictions. (b) despite being heavy in model sizes, Baichuan-2 7B model achieves better performances than the encoder-based models on all the subtasks. The clear advantage of generative models is that they make predictions with the LM prediction head, so it does not require any additional architectural design for different subtasks. All we need to do is to formulate the task into prompt-response pairs. (c) Among the encoder-based methods, we can see that the biaffine based method, UNIRE, outperforms the Sep-Biaffine method, showing the adavantage of unifying entity detection and relation classification into a unified space. (d) COT style generation helps the LLMs to achieve better performances on all three sub-tasks, showing that COT is also helpful under manually designed steps for task solving and fine-tuning. This observation is consistent with PromptCBLUE \cite{PromptCBLUE}. 

\begin{table}[tb!]
\centering
\resizebox{0.99\textwidth}{!}{
\begin{tabular}{cccccccc}

\hline
 \textbf{Subtask}   &    \multicolumn{3}{c}{\textbf{Triplet extract}}    &   \textbf{Node Grouping}    &    \multicolumn{3}{c}{\textbf{Tree assembling}}       \\ 
 \textbf{Metric}   &   Prec   &  Rec   &  F1   &  
 $\text{NG\_LR}$     &  TreeAcc  & DP-F1   &   Tree\_LR     \\
\hline

\multicolumn{8}{c}{\emph{Encoder-based methods}}        \\ 
\hline
UNIRE   &    0.913  &  0.881  &    0.896    &       \\
TPinker   &   0.909   & 0.878    &  0.893    &         \\
CasRel    &    0.882   &  0.891   &  0.886     &     \\
Sep-Biaffine   &    0.893   &  0.897   &   0.895     &      \\ 

NG-Biaffine   &        &      &     &    0.962    &   \\
NG-TableFilling    &       &        &      &     0.961    &    \\ 

TreeAssemble-Biaffine   &      &      &      &        &    0.735
 &    0.841   &   0.937              \\
TreeAssemble-TableFilling    &      &      &      &        &    0.741     &   0.838   &   0.933    \\

\hline
\multicolumn{8}{c}{\emph{Generation-based methods}}        \\ 
\hline
Generation   &   0.901   &  0.894  &  0.897   
  &   0.965   &        0.745    &   0.848   & 
 0.943   \\
COT-Generation   &  0.898   &  0.904  &   0.901    &  0.968    &   0.748   &   0.852    &    0.947      \\

\hline
\end{tabular}}
\caption{\label{tab:pipeline_framework_results}Results for each subtask of the pipeline framework, and the overall result of the Text2MDT task when applying the framework. The average results in five different runs are reported. The best results are in bold. }
\end{table}

\subsubsection{Performances on whole task} \quad Now we consider the following combinations of methods for the complete pipelined predictions of test samples from a given medical text: (a) The encoder based pipeline method (denoted as Enc-Pipe). In Enc-Pipe, UNIRE is responsible for triplet extraction, NG-Biaffine for node grouping, and TreeAssemble-Biaffine for tree assembling. (b) COT-Generation for all the three subtasks (denoted as CGen-Pipe). These two approaches are compared with the five approaches of the end2end framework in Table \ref{tab:end2end_results}. 

From Table \ref{tab:end2end_results}, we can see that: (a) the CGen-Pipe achieves better performances than the Enc-Pipe method, which is natural since COT-Generation performs better than the encoder based models on all three subtasks. (b) Interestingly, the pipelined method CGen-Pipe performs better than the direct Generation method, but does not perform better than COT-Generation-3. Intuitively, the pipelined method CGen-Pipe suffer from error propagation from different steps in the pipeline, thus, although the pipelined methods utilize more model parameters, it does not perform better than the end2end methods. (c) The COT style generation methods perform better than the direct Generation method, which is intuitively sound. Our Text2MDT task is a complex information extraction task containing multiple steps. The COT based generative methods injects priors on how the models should solve the task, thus it can be more informed to use the results of the previously generated contents for future tokens' generation. (d) Intuitively, the generative LMs should benefit more from more detailed and fine-grained COT instructions. However, Table \ref{tab:end2end_results} shows that COT-Generation-3 performs the best. COT-Generation-3 asks the LLMs to first extract triples and then the MDTs, which has fewer thought steps than the other COT based generation methods. We believe that this is because triplet extraction and tree assembling are the two more challenging steps in the three subtasks, and COT-Generation-3 focus itself on these two subtasks during fine-tuning. And COT-Generation-3's thought steps have relatively smaller response length, which is helpful for the LMs to keep track of the generation contents.

\begin{table}[tb!]
\centering
\resizebox{0.6\textwidth}{!}{
\begin{tabular}{cccc}
\hline
 \textbf{Method}   &     TreeAcc  & DP-F1   &   Tree\_ER       \\
\hline
\multicolumn{4}{c}{\emph{Pipeline methods}}        \\ 
\hline
Enc-Pipe   &     0.450   &  0.612    &   0.884          \\
CGen-Pipe   &     0.470  &  0.631    &    0.897   \\
\hline
\multicolumn{4}{c}{\emph{End2end methods}}        \\ 
\hline
Generation   &    0.440   &  0.619  &   0.885  \\
COT-Generation-1    &    0.470   &   0.628    &   0.894  \\
COT-Generation-2    &  0.450   &  0.623   &  0.889  \\  
COT-Generation-3    &   \textbf{0.490}  &   \textbf{0.632}   &   \textbf{0.898}     \\
COT-Generation-4    &   0.450   &  0.626   &  0.892   \\ 

\hline
\end{tabular}}
\caption{\label{tab:end2end_results}Overall results of the pipeline framework and the end2end methods. The average results in five different runs are reported. The best results are in bold.  }
\end{table}



\subsection{Discussions and further analysis}

\subsubsection{Impact of tree depth} \quad In Table \ref{tab:tree_depth_stats}, we present the statistics of the Text2MDT datasets, showing that in our task, medical decision trees are of different depth. In table \ref{tab:results_different_depths}, we present the results of CGen-Pipe, and COT-Generation-3 on different MDT depths. We can see that the two methods obtain the same performance metrics on the MDTs with depth 2. The performance difference between the end2end models and the pipeline methods mainly lie in MDTs with higher depth. And we can see that the performances on the MDTs with depth larger than 2 are significantly worse than those on the MDTs of depth 2. Intuitively, tree depth is a direct reflection of the complexity of treatment procedures for different diseases, and it is expected that test samples with deeper MDTs will present higher difficulty for the models, especially for the generation based models.

\begin{table}[tb!]
\centering
\resizebox{0.98\textwidth}{!}{
\begin{tabular}{c|cccccc}
\hline
\multirow{2}{*}{MDT depth}   &  \multicolumn{3}{c}{CGen-Pipe} &      \multicolumn{3}{c}{COT-Generation-3}     \\ 
 &   TreeAcc  &   DP-F1    &    Tree\_ER       &   TreeAcc  &   DP-F1    &    Tree\_ER     \\ 
\hline
2    &   0.750  &  0.833  &   0.943   &    0.750   &  0.833  &  0.943   \\ 
3    &   0.428   &  0.607  &  0.884    &     0.442  &   0.603  &    0.883   \\ 
4    &   0.454  &  0.648  &   0.946   &     0.545  &   0.676   &  0.950   \\ 
\hline
\end{tabular}}
\caption{\label{tab:results_different_depths}Model performance on medical decision trees of different depths. }
\end{table}

\subsubsection{Impact of backbone models} \quad We first examine how different backbone models affect the Enc-Pipe method. Our main experiments (Table \ref{tab:pipeline_framework_results} and \ref{tab:end2end_results}) use the MedBERT \cite{guo-etal-2021-global} as the backbone. And now we consider the three alternative encoder backbones mentioned in Section \ref{subsec:implement_details}. In the main experiments, we use the Baichuan-2 7B as the generative LM backbone in Table \ref{tab:end2end_results}. Now we consider the nine generative LMs of different scales (presented in Section \ref{subsec:implement_details}) for the COT-Generation-3 method. For the methods with size smaller than 1B, we will fine-tune all the parameters of the backbone. For the larger models, we will fine-tune them with the LoRA method \cite{hu2021lora} (rank = 24). The other experimental settings are kept the same with Section \ref{subsec:implement_details}. 

Table \ref{tab:results_different_backbones} reports the experimental results for different backbone models, and the following observations can be made: (a) for the Enc-Pipe method, the in-domain pre-trained model, MedBERT performs the best among the four pre-trained encoders, showing that further pretraining on the large-scale medical corpus are benefical for the Text2MDT task. This observation is in line with \cite{zhu-2021-mvp,guo-etal-2021-global,Zhu2023ACFAC}. (b) For the generative LMs, models with small parameter sizes performs unsatisfying in our task. The small-scale generative LMs does not have enough language understanding and completion capabilities to face the challenges brought by this task. (c) Among the open-sourced generative LMs we experiment, the Baichuan2 models perform the best, which we believe results from their large-scale pretraining and complete instruction alignment pipeline.

\begin{table}[tb!]
\centering
\resizebox{0.66\textwidth}{!}{
\begin{tabular}{c|ccc}

\hline
 \textbf{Backbone}   &     TreeAcc  & DP-F1   &   Tree\_ER       \\
\hline

\multicolumn{4}{c}{\emph{The Enc-Pipe method}}        \\ 
\hline
MedBERT   &     0.450   &  0.612    &   0.884      \\
BERT-www-ext   &     0.440   &   0.615  &    0.882     \\
BERT-base Chinese  &    0.390   &   0.583   &  0.867     \\
Erlangshen-ZEN1-224M   &   0.410   &   0.596   &   0.873   \\

\hline
\multicolumn{4}{c}{\emph{The COT-Generation-3 method}}        \\ 
\hline
GPT-2 base Chinese    &    0.030   &  0.121   &   0.238  \\
Randeng-T5-784M   &      0.080   &   0.253   &   0.352     \\
BLOOMZ-7.1B-mt    &      0.330   &  0.536   &   0.782    \\
ChatGLM-6B-2    &    0.380   &   0.592   &   0.849  \\
ChatMed    &    0.420    &   0.596   &  0.864    \\
Chinese-LlaMA-2 7B    &     0.410    &    0.581   &  0.868          \\
Chinese-LlaMA-2 13B    &     0.460   &  0.623   &   0.890             \\
Ziya-13B-medical   &     0.450   &   0.614   &  0.886    \\

Baichuan2 7B    &   \textbf{0.490}  &   \textbf{0.632}   &   \textbf{0.898}     \\ 

Baichuan2 13B    &    0.490    &  0.628   &   0.896    \\

\hline
\end{tabular}}
\caption{\label{tab:results_different_backbones}The effects of the pre-trained backbones on the Enc-Pipe and COT-Generation-3 methods. }
\end{table}

\subsubsection{Performance breakdown of the end2end framework}

In Table \ref{tab:perf_breakdown}, we present the performance breakdown of the test predictions by the best performing method in our main experiments, COT-Generation-3. As a comparison, we also present the performance scores of the CGen-Pipe method's final test score on each sub-task. We present how these methods perform on each subtask. Note that with error propagation in the predicted samples, the scores on the node grouping and tree assembling tasks contain errors from the previous subtasks. From Table \ref{tab:perf_breakdown}, we can see that: (a) although the CGen-Pipe method has advantages in the triplet extraction step, the COT-Generation-3 method performs better at the node grouping task, with its generated triplet results. Thus, these two methods already has almost the same score on the node grouping step. And at the tree assembling task, COT-Generation-3 outperforms CGen-Pipe at all three metrics.

\begin{table}[tb!]
\centering
\resizebox{0.72\textwidth}{!}{
\begin{tabular}{cccc}

\hline
 \textbf{Subtask}   &     Metric  &    COT-Generation-3     &   CGen-Pipe     \\
\hline
\multirow{3}{*}{Triplet Extraction}   &     Prec  &     0.894      &   0.897    \\
     &     Rec     &   0.885    &   0.888   \\
     &     F1      &    0.889     &   0.892 \\
\hline

Node grouping     &     NG\_LR    &    0.887   
 &   0.887      \\
\hline

\multirow{3}{*}{Tree assembling}    &    TreeAcc        &    0.490     &   0.47   \\   
&   DP-F1     &     0.632    &   0.631  \\
&   Tree\_LR       &   0.898    &  0.897 \\

\hline
\end{tabular}}
\caption{\label{tab:perf_breakdown}Performance breakdown of the end2end method , COT-Generation-3, and the generation based pipeline method, CGen-Pipe. }
\end{table}

\begin{figure*}[h]
\begin{center}
\includegraphics[width=0.98\textwidth]{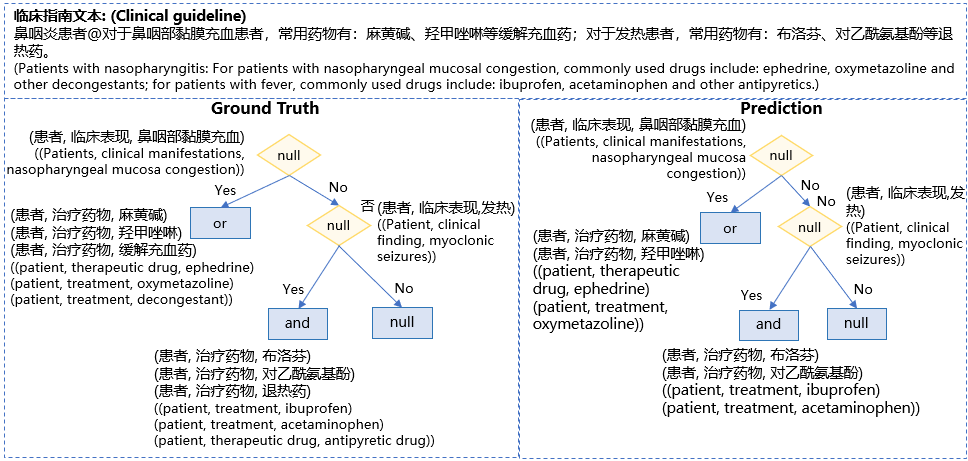}
\end{center}
\caption{Example (a), an error case of the COT-Generation-3 method on the Text2MDT test samples. }
\label{fig:text2dt_case_1}
\end{figure*}

\begin{figure*}[h]
\begin{center}
\includegraphics[width=0.98\textwidth]{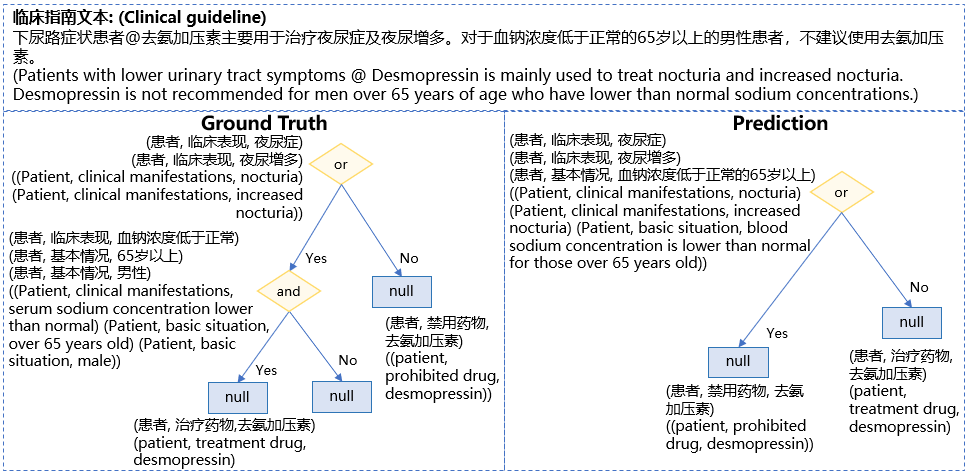}
\end{center}
\caption{Example (b), an error case of the COT-Generation-3 method on the Text2MDT test samples. }
\label{fig:text2dt_case_2}
\end{figure*}

\subsubsection{Case studies}

On the test set of the Text2MDT task, COT-Generation-3 achieves the best performance. Figure \ref{fig:text2dt_case_1} and \ref{fig:text2dt_case_2} report two examples where COT-Generation-3 can not predict the same MDTs with the ground truth. In Figure \ref{fig:text2dt_case_1}, COT-Generation-3 misses the triplet (患者, 治疗药物, 缓解充血药) ((patient, treatment, decongestant)) in the second node, and the triplet (患者, 治疗药物, 退热药) ((patient, therapeutic drug, antipyretic drug)) in the fourth node, during prediciton. These errors are mainly from the triplet extraction subtask, which is the first step of tackling MDTs. In Figure \ref{fig:text2dt_case_2}, COT-Generation-3 made an error in triplet extraction regarding the basic status of the patients, and as a result, made a mistake in node grouping.


\subsection{Limitations}
Our work is the first exploration of extracting MDTs from the medical texts, and our work is currently applicable to some simple scenarios, specifically: 1) The logic expression of nodes is limited. The triplets between nodes are only "and" and "or," while in more complex scenarios, there should be a combination of multiple logical relationships; 2) The expressiveness of the tree is limited. Our decision tree aborts after reaching a decision. The actual scenario should be a process of continuous judgment and decision-making. 3) The length of the text is limited. We only contend to extract one paragraph of medical text; in fact, much medical knowledge needs to be based on multiple sections or even chapters. We will improve it in our future work.

\section{Conclusion}

In this study, we propose a novel task, Text2MDT, which aims to automatically extract medical decision trees from medical texts that are significant for intelligent medicine. We constructed the first Text-to-MDT dataset in the NLP community with medical experts' participation. Since there are no existing neural network based methods that can directly deal with our novel tasks, we propose two cohorts of methods: (a) the pipeline based method, which decomposes the Text2MDT task into three subtasks and utilize the existing methods to complete the subtasks; (b) the end2end method, which is challenging and can not be handled by the encoder-based models. We utilize the recent open-sourced LLMs and chain-of-thought prompting for the end2end methods. Experiments show that: (a) the LLMs can achieve promising results on the Text2MDT benchmark in an end2end fashion. (b) the encoder-based pipeline methods can achieve comparable results with the LLMs while requiring less computations.

\section{Declaration of generative AI in scientific writing}

The authors of this paper hereby declare that no generative artificial intelligence (AI) systems were employed in the production, composition, or generation of the content presented in this scientific work. The manuscript's creation did not involve the utilization of any generative AI tools or systems to produce text, develop ideas, or enhance the overall quality of the content.

The writing process and content refinement were conducted solely by human effort, with the primary aid of conventional writing and proofreading tools such as Grammarly for grammatical accuracy and language refinement purposes. The authors affirm that Grammarly was employed solely as an assistive tool for grammar and language checks and did not contribute to the conceptualization, generation, or development of the scientific ideas presented in this paper.

\section{Author contributions}

We now state the contributions of each author: 
\begin{itemize}
\item Conceptualization: Wei Zhu, Wenfeng Li.
\item Methodology: Wei Zhu, Wenfeng Li, Xing Tian, Pengfei Wang.
\item Resources: Yuan Ni and Guotong Xie.
\item Supervision: Xiaoling Wang, Yuan Ni and Guotong Xie.
\item Validation: Xiaoling Wang, Yuan Ni and Guotong Xie.
\item Writing - original draft: Wei Zhu, Wenfeng Li, Xing Tian, Pengfei Wang.
\item Writing - review \& editing: Xiaoling Wang, Jin Chen, Yuanbin Wu, Yuan Ni and Guotong Xie.
\end{itemize}



\bibliographystyle{elsarticle-harv} 
\bibliography{reference,refs}


\end{CJK*}

\end{document}


\begin{CJK*}{UTF8}{gbsn}

\maketitle

\appendix

\section{Appendix}

\subsection{Prompt and target formats for PromptCBLUE}
\label{subsec:format}

We now present one example of the prompt and response formats for each sub-task of PromptCBLUE. For complete set of prompt templates, please refer to our project page \url{https://github.com/michael-wzhu/PromptCBLUE}.

\noindent\textbf{CMeEE-v2} \quad As a standard named entity recognition task in the medical domain, when given a input text sequence, one has to extract the medical entities, including the text mentions’ spans and entity types, in the sequence. Recently, with the prevailing of generative LM, a series of literature no longer requires the model to predict the entity spans. Thus, we formulate CMeEE-v2 task as the task of generating all the entity mentions. With the idea of \cite{Wei2022ChainOT}, the prompt will ask the LLMs to generate entities type-by-type. In comparison, without COT, the LLMs are asked to generate entities one by one. 

Prompt Template with COT:
\begin{verbatim} 
医学实体识别: (Medical named entity recognition: )
[sentence]
实体选项: [entity_types] (Choices of entity types: [entity_types])
要求：请按照实体类别分别输出实体 (Instruction: please generate
named entities type-by-type. )
答: (Answer: )
\end{verbatim}

Prompt Template without COT:
\begin{verbatim} 
医学实体识别: (Medical named entity recognition: )
[sentence]
实体选项: [entity_types] (Choices of entity types: [entity_types])
答: (Answer: )
\end{verbatim}

Target Template with COT:
\begin{verbatim} 
上述句子中的实体包含: (The above sentence have the following entities: )
[entity\_type]实体: [entity_mentions] ([entity_type] entities: 
[entity_mentions])
\end{verbatim}

Target Template without COT:
\begin{verbatim} 
上述句子中的实体包含: (The above sentence have the following entities: )
[entity\_mention]:[entity\_type]实体
[entity\_mention]:[entity\_type]实体
\end{verbatim}

Example prompt with COT: 
\begin{verbatim}
医学实体识别: (Medical named entity recognition: )
外周血白细胞计数常明显升高，伴核左移。 (The peripheral blood white blood cell 
count is often significantly elevated, with left nuclear shift.)
实体选项: 疾病，医学检验项目，医院科室，身体部位，微生物类，临床表现，药物 (Choices of entity types: diseases, medical test items, hospital 
departments, body parts, microorganisms, clinical manifestations, 
drugs)
要求：请按照实体类别分别输出实体 (Instruction: please generate 
named entities type-by-type. )
答： (Answer: )
\end{verbatim}

Example prompt without COT: 
\begin{verbatim}
医学实体识别: (Medical named entity recognition: )
外周血白细胞计数常明显升高，伴核左移。 (The peripheral blood white blood cell 
count is often significantly elevated, with left nuclear shift.)
实体选项: 疾病，医学检验项目，医院科室，身体部位，微生物类，临床表现，药物 (Choices of entity types: diseases, medical test items, hospital 
departments, body parts, microorganisms, clinical manifestations, 
drugs)
答： (Answer: )
\end{verbatim}

Example target with COT: 
\begin{verbatim} 
上述句子中的实体包含: (The above sentence have the following entities: )
医学检验项目实体: 外周血白细胞计数 (medical test items entities: 
peripheral blood white blood cell count)
疾病实体: 无 (disease entities: None)
医院科室实体: 无 (hospital department entities: None)
药物实体: 无 (drug entities: None)
微生物类实体: 无 (microorganism entities: None)
\end{verbatim}

Example target without COT: 
\begin{verbatim} 
上述句子中的实体包含: (The above sentence have the following entities: )
外周血白细胞计数:医学检验项目实体
\end{verbatim}

Note that we also want LLM to determine whether there are entities for a given type in the sentence.

\noindent\textbf{CMeIE} \quad The CMeIE task is a triple extraction task, that is, the model has to predict the head entity information (including mention span and entity type), tail entity information (including mention
span and entity type), and the relation type between the two entities. Similar as the CMeEE task, we reformuate this task as the task of generating the head-tail entity pair under the given relation types. The following are the prompt and target templates, in which [relations] denotes the relation names, and [sentence] denotes the input text. An example pair of prompt and target response is also presented. 

Prompt Template:
\begin{verbatim} 
找出句子中的具有[relations]的头尾实体对: (Find the entity pairs that 
have the [relations] relations: )
[sentence]
\end{verbatim}

Target Template:
\begin{verbatim} 
上述句子中实体对及其关系如下: (the entity pairs and their relations are:)
[triple] 
[triple] 
\end{verbatim}

Example prompt: 
\begin{verbatim}
找出句子中的具有临床表现，同义词关系类型的头尾实体对： (Find the entity pairs that have the clinical finding or 
synonym relations:)
妊娠期高血压@健康孕妇可以通过全身血管阻力 (systemic vascular resistance，
SVR) 和心输出量下降得以平衡，而妊娠期高血压妇女SVR较低，心输出量往往会增加。 (Hypertension during pregnancy@healthy pregnant women can be 
balanced by a decrease in systemic vascular resistance (SVR) and 
cardiac output, whereas women with gestational hypertension have a 
lower SVR and tend to have an increase in cardiac output.)
\end{verbatim}

Example target: 
\begin{verbatim} 
上述句子中实体对及其关系如下： (the entity pairs and their relations are 
as follows: )
头实体：妊娠期高血压，尾实体：SVR较低；关系：临床表现 (head entity: hypertension during pregnancy, tail entity: SVR 
is lower; relation: clinical finding)
头实体：妊娠期高血压，尾实体：心输出量往往会增加；关系：临床表现 (head entity: hypertension during pregnancy, tail entity: SVR 
is lower; head entity: hypertension during pregnancy, tail entity: 
cardiac output tends to increase; relation: clinical finding)
\end{verbatim}

With the idea of COT, LLMs will deal with this task with a more suitable strategy: first predict the relations in a given sentence, and then extract the triples type by type. 


Prompt Template with COT:
\begin{verbatim} 
找出句子中的具有[relations]的头尾实体对: (Find the entity pairs that 
have the [relations] relations: )
[sentence]
要求：请先判断句子中是否具有某个实体对关系，如果有，则抽取出具有这个关系的实体对 (Instruction: Please determine whether there is a certain relation 
in the sentence. If so, extract the entity pair with this 
relationship. )
\end{verbatim}

Target Template with COT:
\begin{verbatim}
上述句子中包含如下关系: [existing-relations]
上述句子中[relation]关系的实体对如下: [triples] (the entity pairs having 
the [relation] relation are as follows: [triples])
上述句子中[relation]关系的实体对如下: [triples] (the entity pairs having 
the [relation] relation are as follows: [triples])
\end{verbatim}

Example prompt with COT: 
\begin{verbatim}
找出句子中的具有临床表现，同义词关系类型的头尾实体对： (Find the entity pairs that have the clinical finding or 
synonym relations:)
妊娠期高血压@健康孕妇可以通过全身血管阻力 (systemic vascular resistance，
SVR) 和心输出量下降得以平衡，而妊娠期高血压妇女SVR较低，心输出量往往会增加。 (Hypertension during pregnancy@healthy pregnant women can be 
balanced by a decrease in systemic vascular resistance (SVR) and 
cardiac output, whereas women with gestational hypertension have a 
lower SVR and tend to have an increase in cardiac output.)
要求：请先判断句子中是否具有某个实体对关系，如果有，则抽取出具有这个关系的实体对 (Instruction: Please determine whether there is a certain relation 
in the sentence. If so, extract the entity pair with this 
relationship. )
\end{verbatim}

Example target with COT: 
\begin{verbatim} 
上述句子中包含如下关系: 临床表现关系
上述句子中临床表现关系的实体对如下： (the entity pairs having 
the clinical finding relation are as follows: )
头实体：妊娠期高血压，尾实体：SVR较低；头实体：妊娠期高血压，尾实体：心输出量往往会增加； (head entity: hypertension during pregnancy, tail entity: SVR 
is lower; head entity: hypertension during pregnancy, tail entity: 
cardiac output tends to increase;)
\end{verbatim}

\noindent\textbf{CHIP-CDEE} \quad In this task, one needs to extract clinical findings as medical events in a given medical report. An clinical finding event consists of a 主体词 (trigger), and three attributes, 发生状态 (occurrence status), 描述词 (descriptor) and 解剖部位 (
anatomical part). In PromptCBLUE, this task is reformulated as generating event descriptions for clinical findings given a medical report. 

Prompt Template:
\begin{verbatim} 
[sentence]
问题：句子中的临床发现事件及其属性是什么？ (Question: What are the clinical 
findings and their attributes in the sentence?)
说明：临床发现事件由主体词，发生状态，描述词和解剖部位组成 (Note: a clinical finding event consists of trigger, occurrence 
status, descriptor and anatomical part.)
\end{verbatim}

Target Template:
\begin{verbatim} 
上述句子中的临床发现事件如下： (The clinical finding events in the above 
sentence are as follows:)
主体词：[str]；发生状态：[str]；描述词：[str]；解剖部位：[str] (trigger: [str]; occurrence status: [str]; descriptor: 
[str]; anatomical part: [str])
\end{verbatim}

Example prompt: 
\begin{verbatim}
7月前患者给予亚砷酸氯化钠(伊泰达)注射液 10mg 静滴14天，6月前予以口服维甲酸 20mg bid*14天维持治疗，5月前行亚砷酸氯化钠(伊泰达)注射液 10mg 静滴14天维持化疗，3月余前复查骨髓检查示增生性骨髓象；fish：pml/rara（双色双融合）(15/17)：未见异常；腰穿脑脊液未见异常细胞。现为维持化疗入院。(Before July, patients were given 10mg of sodium arsenite chloride 
(Itada) injection intravenously for 14 days. Before June, they were 
given oral retinoic acid 20mg bid * 14 days for maintenance treatment. 
Before May, patients were given 10mg of sodium arsenite chloride 
(Itada) injection intravenously for 14 days for maintenance 
chemotherapy. After more than 3 months, bone marrow examination 
showed proliferative myelogram; Fish: pml/rara (dual color dual 
fusion) (15/17): no abnormalities found; No abnormal cells were 
found in the cerebrospinal fluid through lumbar puncture. I am 
currently admitted for maintenance chemotherapy.)
问题：句子中的临床发现事件及其属性是什么？ (Question: What are the clinical 
findings and their attributes in the sentence?)
说明：临床发现事件由主体词，发生状态，描述词和解剖部位组成 (Note: a clinical finding event consists of trigger, occurrence 
status, descriptor and anatomical part.)
\end{verbatim}

Example target: 
\begin{verbatim}
上述句子中的临床发现事件如下： (The clinical finding events in the above 
sentence are as follows:)
主体词：fish：pml/rara（双色双融合）(15/17)异常；发生状态：否定；描述词：无；解剖部位：无 (trigger: Fish: pml/rara (dual color dual fusion) (15/17) anomaly; 
occurrence status: not found; descriptor: none; anatomical part: none)
主体词：骨髓象；发生状态：无；描述词：增生性；解剖部位：骨髓 (trigger: bone marrow imaging; occurrence status: none; descriptor: 
proliferative; anatomic site: bone marrow)
\end{verbatim}

With COT, we ask the LLMs to conduct event extraction step-by-step. That is ,we first ask the LLMs to detect trigger words, and then we will extract the attributes for each trigger word to complete the events.

Prompt Template with COT:
\begin{verbatim} 
[sentence]
问题：句子中的临床发现事件及其属性是什么？ (Question: What are the clinical 
findings and their attributes in the sentence?)
说明：临床发现事件由主体词，发生状态，描述词和解剖部位组成 (Note: a clinical finding event consists of trigger, occurrence 
status, descriptor and anatomical part.)
请先抽取临床发现事件的主体词，然后为每个主体词补充发生状态，描述词和解剖部位属性。
\end{verbatim}

Target Template with COT:
\begin{verbatim} 
上述句子中的临床发现事件的主体词如下： (The trigger words of the clinical 
finding events in the above sentence are as follows:)
[triggers]；
主体词[trigger]的属性如下：发生状态：[str]；描述词：[str]；解剖部位：[str] (the attributes of the trigger word [trigger]: [str]; 
occurrence status: [str]; descriptor: [str]; anatomical part: [str])
\end{verbatim}

Example prompt with COT: 
\begin{verbatim}
7月前患者给予亚砷酸氯化钠(伊泰达)注射液 10mg 静滴14天，6月前予以口服维甲酸 20mg bid*14天维持治疗，5月前行亚砷酸氯化钠(伊泰达)注射液 10mg 静滴14天维持化疗，3月余前复查骨髓检查示增生性骨髓象；fish：pml/rara（双色双融合）(15/17)：未见异常；腰穿脑脊液未见异常细胞。现为维持化疗入院。(Before July, patients were given 10mg of sodium arsenite chloride 
(Itada) injection intravenously for 14 days. Before June, they were 
given oral retinoic acid 20mg bid * 14 days for maintenance treatment. 
Before May, patients were given 10mg of sodium arsenite chloride 
(Itada) injection intravenously for 14 days for maintenance 
chemotherapy. After more than 3 months, bone marrow examination 
showed proliferative myelogram; Fish: pml/rara (dual color dual 
fusion) (15/17): no abnormalities found; No abnormal cells were 
found in the cerebrospinal fluid through lumbar puncture. I am 
currently admitted for maintenance chemotherapy.)
问题：句子中的临床发现事件及其属性是什么？ (Question: What are the clinical 
findings and their attributes in the sentence?)
说明：临床发现事件由主体词，发生状态，描述词和解剖部位组成 (Note: a clinical finding event consists of trigger, occurrence 
status, descriptor and anatomical part.)
请先抽取临床发现事件的主体词，然后为每个主体词补充发生状态，描述词和解剖部位属性。
\end{verbatim}

Example target with COT: 
\begin{verbatim}
上述句子中的临床发现事件的主体词如下： (The trigger words of the clinical 
finding events in the above sentence are as follows:)
fish：pml/rara（双色双融合）(15/17)异常；骨髓象
主体词\"pml/rara（双色双融合）(15/17)异常\"的属性如下：发生状态：否定；描述词：无；解剖部位：无 (trigger word \"Fish: pml/rara (dual color dual fusion) (15/17) 
anomaly\" has the following attributes: occurrence status: not 
found; descriptor: none; anatomical part: none)
主体词\"骨髓象\"的属性如下：发生状态：无；描述词：增生性；解剖部位：骨髓 (trigger: bone marrow imaginghas the following attributes: 
occurrence status: none; descriptor: proliferative; anatomic site: 
bone marrow)
\end{verbatim}

\noindent\textbf{CHIP-CDN} \quad This task aims to map the diagnosis terms (query terms) written by doctors to standardized disease terms (target terms) according to a certain medical terminology standard \cite{jmir}. The number of standardized terms may exceed 10 thousands. The standard system adopted in this task is usually the International Statistical Classification of Diseases and Related Health Problems 10th Revision (ICD-10)\footnote{https://www.who.int/standards/classifications/classification-of-diseases}. ICD-10 has more than 30 thousand disease terms, thus it is prohibitive to feed all the disease terms into ChatGPT. ICD-10 has different versions in different countries, and in this work we adopt the ICD-10 Beijing Clinical Trial (version v601)\footnote{http://www.cips-chip.org.cn/2021/eval3}. We will refer to this Chinese version of ICD-10 as ICD-10-Beijing. 

Recently, this task is usually addressed by a retrieve-and-rank system \cite{Park2019AnIR}. Firstly, a small set of candidate standard terms are retrieved via inverted index (e.g., ElasticSearch) or a semantic index toolkit like Faiss, then a ranking model will score and rank the relevances of each query-target term pair. To construct the data samples in CBLUE, for each query term in the task datasets, we first retrieve candidate target terms using BM25 \cite{Robertson2009ThePR}. The candidate pool consists of the top 10-30 terms retrieved by BM25 not are not in the ground truth, and the ground truth terms (for 80\% of the cases).  The candidate pool's order is shuffled so that LLMs can not obtain the true answers by just selecting the first few candidates. Thus, under PromptCBLUE, LLMs act as a ranking model, and choose the final target terms among the candidates or determine no proper target terms are present in the candidate pool.

Prompt Template:
\begin{verbatim} 
[sentence]
术语选项：[candidate-terms] (choices of terms：[candidate-terms])
说明：从候选的若干个ICD-10诊断标准词中选择出与原诊断描述匹配的词 (Task explanation: select the terms that match the original 
diagnostic description from several candidate ICD-10 diagnostic 
standard terms)
答： (Answer)
\end{verbatim}

Target template:
\begin{verbatim} 
[selected-terms]
\end{verbatim}

Example prompt: 
\begin{verbatim}
主动脉弓缩窄心功能低下 (Aortic arch constriction and cardiac dysfunction)
术语选项：胫前动脉假性动脉瘤，主动脉缩窄，男性性腺功能低下，男性性腺功能低下，垂体功能低下，心功能不全 (choices of terms：pseudoaneurysm of the anterior tibial artery, 
coarctation of the aorta, male hypogonadism, male hypogonadism, 
pituitary dysfunction, and cardiac insufficiency)
说明：从候选的若干个ICD-10诊断标准词中选择出与原诊断描述匹配的词 (Task explanation: select the terms that match the original 
diagnostic description from several candidate ICD-10 diagnostic 
standard terms)
答： (Answer)
\end{verbatim}

Example target: 
\begin{verbatim}
主动脉缩窄，心功能不全 (coarctation of the aorta, cardiac insufficiency)
\end{verbatim}

\noindent\textbf{CHIP-CTC} \quad The main objective of this task is to classify the text input into a type of clinical trial screening criteria (44 types). Under PromptCBLUE, we transform the task to the task of generate the type of clinical trial screening criteria for the give text input and the given label set. LLMs need to respond  with "Not the above type" to indicate that the given label types in the prompt does not match the sentence.

Prompt template:
\begin{verbatim} 
[sentence]
这句话是什么临床试验筛选标准类型？ (What is the type of clinical trial 
screening criteria for this sentence?)
类型选项：[candidate-types] (Choices of categories: [candidate-types])
\end{verbatim}

Target template:
\begin{verbatim} 
[selected-type]
\end{verbatim}

Example prompt: 
\begin{verbatim}
8.过去3个月内有过眼内手术的患者； (8. Patients who have undergone intraocular 
surgery within the past 3 months;)
这句话是什么临床试验筛选标准类型？ (What is the type of clinical trial 
screening criteria for this sentence?)
类型选项：成瘾行为，吸烟状况，性取向，残疾群体，读写能力，肿瘤进展，参与其它试验，疾病分期，能力，疾病，药物，诊断，教育情况，口腔相关，受体状态，健康群体，数据可及性，设备，献血，过敏耐受，特殊病人特征，睡眠，怀孕相关，研究者决定，器官组织状态，症状(患者感受)，治疗或手术，护理，性别，种族，实验室检查，知情同意，饮食，年龄，居住情况，病例来源，酒精使用，体征(医生检测），锻炼，风险评估，预期寿命，伦理审查，依存性 (Choices of categories: addiction behavior, smoking status, sexual 
orientation, disabled population, reading and writing ability, tumor 
progression, participation in other trials, disease staging, ability, 
disease, medication, diagnosis, educational status, oral related, 
receptor status, healthy population, data accessibility, equipment, 
blood donation, allergic tolerance, special patient characteristics, 
sleep, pregnancy related, researcher decision, organ and tissue status, 
symptoms (patient perception), treatment or surgery, Nursing, gender, 
race, laboratory examination, informed consent, diet, age, residential 
status, source of case, alcohol use, physical signs (doctor tested), 
exercise, risk assessment, life expectancy, ethical review, dependency)
\end{verbatim}

Example target: 
\begin{verbatim}
治疗或手术 (treatment or surgery)
\end{verbatim}

\noindent\textbf{KUAKE-QIC} \quad This task asks one to classify the intent labels for an online medical search query. There is a relatively high proportion of samples with the the "other" category. Thus, similar to the KUAKE-QIC task, we drop this label and reformulate this task as generating the intent type of a given search query with the candidate labels given in the prompt, or respond with "Not the above type" to indicate that the given label types in the prompt does not match the sentence.

Prompt Template:
\begin{verbatim} 
判断下面搜索词的意图： (Determine the intention of the following 
search term)
[query] 
选项：[candidate-types] (Choices: [candidate-types])
答： (Answer: )
\end{verbatim}

Target template:
\begin{verbatim} 
[selected-type]
\end{verbatim}

Example prompt: 
\begin{verbatim}
判断下面搜索词的意图： (Determine the intention of the following search 
term)
武汉传染性尖锐湿疣的治疗方法 (Treatment methods for infectious genital 
warts in Wuhan)
选项：指标解读，治疗方案，功效作用，注意事项，病情诊断，就医建议，疾病描述 (Choices: Indicator interpretation, treatment plan, efficacy, 
precautions, condition diagnosis, medical advice, disease description)
答： (Answer: )
\end{verbatim}

Example target: 
\begin{verbatim}
治疗方案 (treatment plan)
\end{verbatim}

\noindent\textbf{CHIP-STS} \quad The aim of this task is to determine whether two disease-related questions express the same semantics. Under PromptCBLUE, the task is to respond 是 (\emph{yes}), or 相同 (the same), or "不是" (\emph{no}), or 不同 (different) to indicate whether the two input medical questions in the prompt have the same semantics.

Prompt Template:
\begin{verbatim} 
下面两个句子语义是“相同”或“不同”？ (Are the following two sentences 
semantically the same or different?)
“[sentence-1]”，“[sentence-2]”。
选项：[candidate-types]
答：(Answer: )
\end{verbatim}

Target template:
\begin{verbatim} 
[selected-type]
\end{verbatim}

Example prompt: 
\begin{verbatim}
下面两个句子语义是“相同”或“不同”？ (Are the following two sentences 
semantically the same or different?)
“糖尿病的三多一少是什么”，“无限极的“灵芝皇”和“桑唐饮”能治好糖尿病吗？”。 ("What is diabetes' three more and one less ", "Can the 
limitless" Ganoderma lucidum emperor "and" Sangtang drink "cure 
diabetes?".)
选项：相同，不同 (Option: the ame, different)
答：(Answer: )
\end{verbatim}

Example target: 
\begin{verbatim}
不同 (different)
\end{verbatim}

\noindent\textbf{KUAKE-QTR} \quad Under CBLUE, this task asks one to determine to what degree the online medical search query and the title of a web page are related semantically by outputing a relevance score. The relevance score is one of the followng four values: 0 (means "完全不匹配", \emph{completely mismatched or without any reference value}), 1 (means "很少匹配有一些参考价值", \emph{slightly matched but having some reference value}), 2 (means "部分匹配", \emph{partially matched}) and 3 (means "完全匹配", \emph{perfectly matched}). Under PromptCBLUE, this task is formulated as outputing the label names indicating to what degree a query and a page title are related. 

Prompt Template:
\begin{verbatim} 
下面的搜索词和页面标签的意思有多相同？ (How do the following search terms 
and page tags have the same meaning?)
搜索词：[query] (Search term: [query])
页面标题：[web-page-title] (Page Title: [web-page-title])
选项：完全不匹配或者没有参考价值，很少匹配有一些参考价值，部分匹配，完全匹配 (Option: Completely mismatched or without reference value, rarely 
matched with some reference value, partially matched, completely 
matched)
\end{verbatim}

Target template:
\begin{verbatim} 
[selected-type]
\end{verbatim}

Example prompt: 
\begin{verbatim}
下面的搜索词和页面标签的意思有多相同？ (How do the following search terms 
and page tags have the same meaning?)
搜索词：宝宝三周了发烧不玩睡觉 (Search term: The baby has been running 
a fever for three weeks, he doesn't play and just sleep)
页面标题：孩子三周了手足口发烧一天就不烧了就是睡觉打搀 (Page Title: The child has been running a fever in his hands, feet, 
and mouth for three weeks, but after a day, he stops burning and 
goes to bed)
选项：完全不匹配或者没有参考价值，很少匹配有一些参考价值，部分匹配，完全匹配 (Option: Completely mismatched or without reference value, rarely 
matched with some reference value, partially matched, completely 
matched)
\end{verbatim}

Example target: 
\begin{verbatim}
部分匹配 (partially matched)
\end{verbatim}

\noindent\textbf{KUAKE-QQR} \quad This task is designated to differentiate the semantic relations between two medical queries. And the semantic relation labels are: -1 (meaning "语义无直接关联", \emph{not semantically related}), 0 (meaning "后者是前者的语义父集", \emph{the latter is the semantic superset of the former}), 1 (meaning "后者是前者的语义子集", \emph{the latter is the semantic subset of the former}), 2 (meaning "完全一致", \emph{completely equivalent}).

Prompt template:
\begin{verbatim} 
下面两个句子的语义关系是？ (What is the semantic relationship between the 
following two sentences?)
“[sentence-1]”，“[sentence-2]”。 
选项: 完全一致，后者是前者的语义子集，后者是前者的语义父集，语义毫无关联 (Option: completely consistent, the latter is a semantic subset 
of the former, and the latter is a semantic superset of the former, 
with no semantic correlation)
\end{verbatim}

Target template:
\begin{verbatim} 
[selected-type]
\end{verbatim}

Example prompt: 
\begin{verbatim}
下面两个句子的语义关系是？ (What is the semantic relationship between 
the following two sentences?)
“伤口涂什么药好得快”，“有伤口涂什么药”。 (What kind of medicine is good 
for wounds quickly? What kind of medicine is good for wounds.) 
选项: 完全一致，后者是前者的语义子集，后者是前者的语义父集，语义毫无关联 (Option: completely consistent, the latter is a semantic subset 
of the former, and the latter is a semantic superset of the former, 
with no semantic correlation)
\end{verbatim}

Example target: 
\begin{verbatim}
完全一致 (completely consistent)
\end{verbatim}

\noindent\textbf{KUAKE-IR} \quad This is originally a standard information retrieval (IR) task in the medical domain. It asks one to retrieve relevant documents for a online medical query among a corpus of 1000 thousand documents. Apparently, one can not input all the documents in the corpus into LLMs. Thus, we reformulate this task to the task of determine whether the given query and document are relevant. 

Prompt template:
\begin{verbatim} 
医疗搜索：[query] (medical search: [query])
回答内容：[document] (content: [document])
选项: 相关，不相关 (Option: related, unrelated)
答： (Answer: )
\end{verbatim}

Target template:
\begin{verbatim} 
[selected-type]
\end{verbatim}

Example prompt: 
\begin{verbatim}
医疗搜索：鼻梁被撞鼻梁矫正手术 (medical search: correction surgery for 
nasal bridge collision)
回答内容：你好，你这中情况一般需要行鼻骨截骨整形及鼻中隔联合矫正，手术需要住院，大概需要10天左右的时间，费用在12000左右，我们医院不对医保，如果是要医保报销需要办转诊手续之后再凭相关单据回所在地报销。 (content: Hello, in this case, you usually need to undergo nasal 
bone osteotomy and nasal septum joint correction. The surgery requires 
hospitalization, which takes about 10 days and costs around 12000 
yuan. Our hospital does not provide medical insurance. If you want 
medical insurance reimbursement, you need to go through referral 
procedures and then return to your location with relevant documents 
for reimbursement.)
选项: 相关，不相关 (Option: related, unrelated)
答： (Answer: )
\end{verbatim}

Example target: 
\begin{verbatim}
相关 (unrelated)
\end{verbatim}

\noindent\textbf{CHIP-MDCFNPC} \quad Given a dialogue history between a patient and a doctor, this task asks one to extract the clinical finding entities and their statuses. Here the status of a clinical finding entity is defined as one of the four categories: "阳性" ("已有症状疾病或者假设未来可能发生的疾病", \emph{existing symptoms/diseases or potential future symptoms/diseases}), "阴性" (meaning "未患有的症状疾病", \emph{symptoms/diseases not currently experienced}), 其他 ("没有回答、不知道、回答不明确或者模棱两可不好推断", \emph{no answer, don't know, unclear answer, or ambiguous response that is not conducive to inference.}), 不标注 ("无实际意义的不标注或者和病人当前的状态独立不标注", \emph{irrelevant and not specified or unrelated to the patient's current condition and not specified}). Under PromptCBLUE, we ask LLMs to output the descriptions of the status label for extracted clinical finding entities.

Prompt template:
\begin{verbatim} 
[dialogue-history]
问题：上述问诊对话中临床发现有哪些？这些实体的阴阳性是？ (Question: What are the clinical findings in the above consultation 
dialogue? What are the statuses of these entities?)
阴阳性选项：已有症状疾病或者假设未来可能发生的疾病等，未患有症状疾病，没有回答、不知道、回答不明确或者模棱两可不好推断，无实际意义的不标注或者和病人当前的状态独立不标注 (Options: Existing symptomatic diseases or assumed future diseases, 
etc., without symptomatic diseases, without answers, unclear answers, 
or ambiguous reasoning, without practical significance, or independent 
of the patient's current state)
说明：临床发现是临床医学下，病人状态描述的概念集合 (Explanation: Clinical findings are a collection of concepts 
describing patient states in clinical medicine)
\end{verbatim}

Target template:
\begin{verbatim} 
上述对话中临床发现实体以及其阴阳性判别如下：(The clinical finding entities detected in the above conversation 
and its status are as follows: )
[clinical-finding-entity]：[status-option]
\end{verbatim}

Example prompt: 
\begin{verbatim}
患者：月经来了还可吃乌鸡白凤丸和丹栀逍遥丸吗 (Patient: Can I still take Wuji Baifeng 
Pills and Danzhi Xiaoyao Pills after menstruation)
医生：请问类似症状出现多长时间？ (Doctor: How long has similar symptoms 
occurred?)
医生：你吃这药是治疗什么的 (Doctor: What is the purpose of taking this 
medicine for)
患者：我前几个月去检查是游离子腺素增高，月经没来 (Patient: I went to check a few months ago for an increase in free 
radicals, and my menstrual cycle did not come)
医生：什么高 (Doctor: what is high?)
患者：甲状游离子腺素增高 (Patient: Increased thyroid free radical hormone)
医生：把化验单给我看一下。 (Doctor: Show me the test report.)
患者：现在去检查正常值了 (Patient: Now we're going to check the normal values)
医生：那你如果月经量多这些药就不吃了，如果月经量少就可以吃。 (Doctor: If you have a lot of menstrual flow, you can skip 
these medications. If you have a little menstrual flow, you can take them.)
患者：就是月经不调 (Patient: It's just menstrual irregularities)
患者：甲状腺素药还有吃 (Patient: Still taking thyroid hormone medication)
医生：是甲状腺功能低下吗？甲减吗？ (Doctor: Is it hypothyroidism? Is 
hypothyroidism present?)
患者：我在马来西亚看不懂报告单 (Patient: I cannot understand the report 
form in Malaysia)
医生：嗯嗯，只有甲状腺功能低下才需要吃甲状腺素。 (Doctor: Hmm, only hypothyroidism requires taking thyroid hormone.)
问题：上述问诊对话中临床发现有哪些？这些实体的阴阳性是？ (Question: What are the clinical findings in the above consultation 
dialogue? What are the statuses of these entities?)
阴阳性选项：已有症状疾病或者假设未来可能发生的疾病等，未患有症状疾病，没有回答、不知道、回答不明确或者模棱两可不好推断，无实际意义的不标注或者和病人当前的状态独立不标注 (Options: Existing symptomatic diseases or assumed future diseases, 
etc., without symptomatic diseases, without answers, unclear answers, 
or ambiguous reasoning, without practical significance, or independent 
of the patient's current state)
说明：临床发现是临床医学下，病人状态描述的概念集合 (Explanation: Clinical findings are a collection of concepts describing 
patient states in clinical medicine)
\end{verbatim}

Example target: 
\begin{verbatim}
上述对话中临床发现实体以及其阴阳性判别如下：(The clinical finding entities detected in the above conversation 
and its status are as follows: )
月经没来：已有症状疾病或者假设未来可能发生的疾病等 (Menstruation not coming: Existing symptomatic diseases or assumed 
future diseases, etc.)
游离子腺素增高：已有症状疾病或者假设未来可能发生的疾病等 (Elevated free radical adenosine: Existing symptomatic diseases 
or assumed future diseases, etc.)
甲状游离子腺素增高：已有症状疾病或者假设未来可能发生的疾病等 (Increased thyroid free radical hormone: Existing symptomatic 
diseases or assumed future diseases, etc.)
月经量少：无实际意义的不标注或者和病人当前的状态独立不标注 (Low menstrual flow: No labeling of meaningless or independent of 
the patient's current state)
月经量多：无实际意义的不标注或者和病人当前的状态独立不标注 (Menorrhagia: Low menstrual flow: No labeling of meaningless or 
independent of the patient's current state)
月经不调：已有症状疾病或者假设未来可能发生的疾病等 (Irregular menstruation: Existing symptomatic diseases or assumed 
future diseases, etc.)
甲减：没有回答、不知道、回答不明确或者模棱两可不好推断 (Hypothyroidism: Not answering, not knowing, unclear or ambiguous 
in answer, making it difficult to infer)
甲状腺功能低下：无实际意义的不标注或者和病人当前的状态独立不标注 (Low menstrual flow: No labeling of meaningless or independent of 
the patient's current state)
\end{verbatim}

\noindent\textbf{IMCS-V2-SR} \quad Under CBLUE, this task consists of multiple steps: (a) extract symptom entities from a dialogue history between a patient and a doctor; (b) normalize the symptom entity mentions to standardized terms (choosing from the vocabulary provided by the task); (c) determine the status of a symptom entity. In this task, the status label set is: "患有该症状" (\emph{experiencing/suffering from the symptom}), "没有患有该症状" (\emph{not experiencing/suffering from that symptom}), "无法根据上下文确定病人是否患有该症状" (unable to determine from the context whether the patient is experiencing that symptom). Under PromptCBLUE, we discard the second step, that is, we only ask the LLMs to extract the symptom entities in the current utterance and determine their statuses.

Prompt template:
\begin{verbatim} 
找出当前对话中的症状，并判断阴阳性： (Identify the symptoms in the current 
conversation and determine their status:)
对话历史：(Dialogue history: )
[dialogue-history]
当前对话：(Current utterance: )
[current-utterance]
症状阴阳性选项：没有患有该症状，患有该症状，无法根据上下文确定病人是否患有该症状 (Options of symptom status: not experiencing/suffering from that 
symptom, experiencing/suffering from the symptom, unable to determine 
from the context whether the patient is experiencing that symptom)
\end{verbatim}

Target template:
\begin{verbatim} 
当前对话中的症状及其阴阳性判断为：(The symptoms in the current conversation 
and their statuses are: )
[aymptom]：[status]
\end{verbatim}

Example prompt: 
\begin{verbatim}
找出当前对话中的症状，并判断阴阳性： (Identify the symptoms in the current 
conversation and determine their status:)
对话历史： (Dialogue history: )
患者：没有怎么听啊 (Patient: I didn't listen very much)
医生：根据您的描述，宝宝咳嗽，嗓子吼，可能是气喘或喉鸣，考虑支气管炎的可能性较大 (Doctor: According to your description, the baby is coughing, 
roaring in the throat, possibly due to wheezing or wheezing in the 
throat, and the possibility of bronchitis is higher)
当前对话： (Current utterance: )
医生：需要带宝宝去医院儿科就诊，用听诊器听诊肺部，查血常规胸片等相关检查，排除肺炎，根据结果，给于控制感染，止咳化痰等对症治疗。 (Doctor: It is necessary to take the baby to the pediatric 
department of the hospital for treatment, use a stethoscope to 
auscultate the lungs, conduct blood routine chest X-ray and other 
related tests, rule out pneumonia, and provide symptomatic treatment 
such as infection control, cough relief, and phlegm reduction 
based on the results.)
症状阴阳性选项：没有患有该症状，患有该症状，无法根据上下文确定病人是否患有该症状 (Options of symptom status: not experiencing/suffering from that 
symptom, experiencing/suffering from the symptom, unable to determine 
from the context whether the patient is experiencing that symptom)
\end{verbatim}

Example target: 
\begin{verbatim}
当前对话中的症状及其阴阳性判断为：(The symptoms in the current conversation 
and their statuses are: )
肺炎：无法根据上下文确定病人是否患有该症状 (Pneumonia: unable to determine from the context whether the patient
is experiencing that symptom)
感染：患有该症状 (Infection: experiencing/suffering from the symptom)
咳：患有该症状 (Cough: experiencing/suffering from the symptom)
痰：患有该症状 (Phlegm: experiencing/suffering from the symptom)
\end{verbatim}

\noindent\textbf{IMCS-V2-NER} \quad The reformulation of this task is similar to that of CMeEE-v2, except that this task focuses on NER of dialogue utterances. 

Prompt template:
\begin{verbatim} 
下面对话中的[entity-labels]实体有哪些？ (What are the [entity-labels] 
entities in the following conversation?)
[utterance]
答： (Answer: )
\end{verbatim}

Target template:
\begin{verbatim} 
上述句子中的实体包含： (The entities in the above sentence include: )
[entity-mention]：[entity-type]实体
[entity-mention]：[entity-type]实体
\end{verbatim}

Example prompt: 
\begin{verbatim}
下面对话中的医学检查检验，症状，医疗操作实体有哪些？ (What are the medical examinations, symptoms, and medical procedures 
in the following conversation?)
宝贝也呕吐吗？ (Does the baby also vomit?)
答： (Answer: )
\end{verbatim}

Example target: 
\begin{verbatim}
上述句子中的实体包含： (The entities in the above sentence include: )
呕吐: 症状实体 (vomiting: Symptom entity)
\end{verbatim}

With the idea of COT \cite{Wei2022ChainOT}, the prompt will ask the LLMs to generate entities type-by-type. 

Prompt template with COT:
\begin{verbatim} 
下面对话中的[entity-labels]实体有哪些？ (What are the [entity-labels] 
entities in the following conversation?)
[utterance]
说明：请根据实体类型进行抽取实体 (Note: Please extract entities according to the entity type)
答： (Answer: )
\end{verbatim}

Target template with COT:
\begin{verbatim} 
上述句子中的实体包含： (The entities in the above sentence include: )
[entity-type]实体：[entity-mentions]
\end{verbatim}

Example prompt with COT: 
\begin{verbatim}
下面对话中的医学检查检验，症状，医疗操作实体有哪些？ (What are the medical examinations, symptoms, and medical procedures 
in the following conversation?)
宝贝也呕吐吗？ (Does the baby also vomit?)
说明：请根据实体类型进行抽取实体 (Note: Please extract entities according to the entity type)
答： (Answer: )
\end{verbatim}

Example target with COT: 
\begin{verbatim}
上述句子中的实体包含： (The entities in the above sentence include: )
医学检查检验实体：无 (Medical examination and testing entity: None)
症状实体：呕吐 (Symptom entity: vomiting)
医疗操作实体：无 (Medical operation entity: None)
\end{verbatim}

\noindent\textbf{IMCS-V2-DAC} \quad This task askes one to classify the intent of the current utterance in a patient-doctor dialogue for online medical consultations. The original task only gives label names that are not natural languages. To better fit this task into LLMs, we re-write the intent label names. 

Prompt template:
\begin{verbatim} 
确定这句话的意图: (Determine the intent label of this utterance: )
[utterance]
类型选项：[candidate-labels] (Choices of intent labels: [candidate-labels])
\end{verbatim}

Target template:
\begin{verbatim} 
[selected-label]
\end{verbatim}

Example prompt: 
\begin{verbatim}
确定这句话的意图: (Determine the intent label of this utterance: )
当时医生说我们单纯支气管炎也不喘就开的药 (At that time, the doctor said we 
prescribed medication for simple bronchitis without wheezing)
类型选项：关于就医建议的解答，给出诊断，关于症状的回答，关于症状的询问，关于就医建议的提问，关于已有检查和治疗的回答，关于注意事项的提问，关于已有检查和治疗的提问，关于个人基本信息的询问，关于个人基本信息的回答，关于用药建议的解答，关于病因的询问，关于用药建议的提问，关于注意事项的解答，关于病因的回答 (Choices of intent labels: Answers to medical advice, providing 
diagnosis, answering symptoms, asking questions about symptoms, 
asking questions about medical advice, answering questions about 
existing tests and treatments, asking questions about precautions, 
asking questions about existing tests and treatments, asking questions 
about personal basic information, answering questions about personal 
basic information, answering questions about medication advice, 
asking questions about causes, and asking questions about medication 
advice, Answers to precautions and etiology)
\end{verbatim}

Example target: 
\begin{verbatim}
关于已有检查和治疗的回答 (answers regarding existing examinations and 
treatments)
\end{verbatim}

\noindent\textbf{IMCS-V2-MRG} \quad In this task, LLMs are asked to summarize a patient-doctor dialogue and write a diagnostic and treatment report. In the dataset, all the reports are divided into 6 sections: chief complaint, present illness history, auxiliary examination, past medical history, diagnosis, recommendations. Thus, under PromptCBLUE, we also ask the LLMs to generate the summarization report in the order of these six sections.

Prompt template:
\begin{verbatim} 
问诊对话历史： (Consultation conversation history: )
[dialogue-history]
根据上述对话，给出诊疗报告 (Based on the above dialogue, a diagnosis and 
treatment report is given)
说明：诊疗报告分为主诉, 现病史, 辅助检查, 既往史, 诊断, 建议这六个章节。 (Note: The diagnosis and treatment report is divided into six 
chapters: chief complaint, history of present illness, auxiliary 
examination, past history, diagnosis, and recommendations.)
\end{verbatim}

Target template:
\begin{verbatim} 
上述问诊对话的诊疗报告如下： (The diagnosis and treatment report of the 
above consultation dialogue is as follows:)
主诉：[str] (Chief complaint: [str])
现病史：[str] (History of current illness: [str])
辅助检查：[str] (Auxiliary check: [str])
既往史：[str] (Past history: [str])
诊断：[str] (Diagnosis: [str])
建议：[str] (Suggestion: [str])
\end{verbatim}

Example prompt: 
\begin{verbatim}
问诊对话历史： (Consultation conversation history: )
患者：宝宝刚满月，母乳喂养，最近两天时不时的会咳嗽一声，食欲和精神还行，只不过睡觉不是很安稳。家里面最近两天大人和宝宝的姐姐也有感冒，不知道宝宝是被传染了感冒还是怎么样，请问怎么治疗？ (Patient: The baby is just one month old and breastfeeding. He 
has coughed from time to time in the past two days. His appetite 
and energy are okay, but his sleep is not very stable. In the past 
two days at home, the adults and the baby's sister have also had colds. 
I don't know if the baby has been infected with a cold or something 
else. How can I treat it?)
医生：您好，我是您的辅诊医生，需要询问几个问题，才能更好的评估孩子情况，您还在吗？ (Doctor: Hello, I am your auxiliary doctor. I need to ask a few 
questions to better evaluate the child's condition. Are you still here?)
医生：宝宝体温正常吗？ (Doctor: Is the baby’s temperature normal?)
医生：还在吗？ (Doctor: Are you still there?)
患者：在 (Patient: in)
医生：您好 (Doctor: Hello)
医生：宝宝现在体温正常吗 (Doctor: Is the baby’s temperature normal now?)
患者：体温正常 (Patient: Temperature is normal)
医生：口吐泡泡吗 (Doctor: Are you spitting bubbles at the mouth?)
患者：没有 (Patient: None)
医生：嗓子哑吗 (Doctor: Do you have a hoarse voice?)
患者：哭起来跟以前一样 (Patient: Crying is the same as before)
医生：好的 (Doctor: OK)
患者：只不过鼻音重 (Patient: It’s just that the nasal sound is heavy)
医生：还有其他症状吗 (Doctor: Are there any other symptoms?)
患者：睡不安稳 (Patient: Unable to sleep well)
医生：出汗多吗？ (Doctor: Do you sweat a lot?)
患者：不多 (Patients: not many)
医生：哭闹吗 (Doctor: Are you crying?)
患者：比以前爱哭闹 (Patient: more crying than before)
医生：大便什么样子 (Doctor: What does the stool look like?)
患者：这个没注意，昨天一天没有大便，今天上午大便的，大便以后睡得安稳一些了，不过还是时不时咳嗽一声 (Patient: I didn’t pay attention to this. He didn’t have a bowel 
movement yesterday. He had a bowel movement this morning. He slept 
more peacefully after the bowel movement, but he still coughed from 
time to time.)
医生：有痰吗? (Doctor: Is there phlegm?)
患者：没有痰，干咳 (Patient: There is no phlegm, and there is dry cough)
医生：嗯嗯 (Doctor: Uh-huh)
医生：流鼻涕吗 (Doctor: Do you have a runny nose?)
患者：没有 (Patient: None)
医生：嗯嗯 (Doctor: Uh-huh)
医生：因为宝宝比较小，最好带宝宝去公立医院儿科就诊，听一下肺部没有问题的话可以观察看看 (Doctor: Because the baby is relatively small, it is best to take 
the baby to the pediatrician in a public hospital. If there is no 
problem with the lungs, he can be observed.)
根据上述对话，给出诊疗报告 (Based on the above dialogue, a diagnosis 
and treatment report is given)
说明：诊疗报告分为主诉, 现病史, 辅助检查, 既往史, 诊断, 建议这六个章节。 (Note: The diagnosis and treatment report is divided into six 
chapters: chief complaint, history of present illness, auxiliary 
examination, past history, diagnosis, and recommendations.)
\end{verbatim}

Example target: 
\begin{verbatim}
上述问诊对话的诊疗报告如下： (The diagnosis and treatment report of the 
above consultation dialogue is as follows:)
主诉：阵发性咳嗽。 (Chief complaint: 
Paroxysmal cough.)
现病史：患儿阵发性干咳两天。 (History of current illness: The child had 
paroxysmal dry cough for two days.)
辅助检查：暂缺。 (Auxiliary check: None)
既往史：不详。 (Past history: None)
诊断：咳嗽待查。 (Diagnosis: Cough to be examined.)
建议：儿科就诊，听诊肺部。 (Suggestion: Pediatric visit, auscultation of 
the lungs.)
\end{verbatim}

\noindent\textbf{MedDG} \quad This task asks one to generate the responses of doctors give the dialogue history and the patient's current utterance. Under the original setting, MedDG is an entity-centric dialogue generation task, that is, it asks the model to generate responses that contains certain entities. Under PromptCBLUE, we currently discard this setting and only asks the LLMs to generate responses on behlf of the doctors. 

Prompt template:
\begin{verbatim} 
[dialogue-history]
根据上述对话历史，作为医生应该如何回复？ (Based on the above conversation 
history, how should a doctor respond?)
答： (Answer: )
\end{verbatim}

Target template:
\begin{verbatim} 
[generated-response]
\end{verbatim}

Example prompt: 
\begin{verbatim}
患者：最近总是到了晚上就胃很难受。这几天吃过饭就有点反胃但是都是头痛头晕恶心。胃还是很难受（女，19岁） (Patient: Recently, I always feel sick in my stomach at 
night. I've been feeling a little nauseous after eating for the past 
few days, but I still have headaches, dizziness, and nausea. My 
stomach still feels uncomfortable (female, 19 years old))
医生：你好，这种情况有多长时间了？ (Doctor: Hello, how long has this been 
going on?)
患者：半个月了。 (Patient: It’s been half a month.)
医生：平时吃饭规律吗？ (Doctor: Do you eat regularly?)
根据上述对话历史，作为医生应该如何回复？ (Based on the above conversation 
history, how should a doctor respond?)
答： (Answer: )
\end{verbatim}

Example target: 
\begin{verbatim}
胃部感觉难受是怎么难受？反酸烧心打嗝？还是胃疼胃胀？ (Why does your stomach feel so bad? Acid reflux, heartburn and 
hiccups? Or stomachache or bloating?)
\end{verbatim}

\subsection{Hyper-parameter settings for the PEFT methods}
\label{subsec:hyperparams}

We now elaborate on the hyper-parameter settings for different PEFT methods, and the training settings.

\textbf{PEFT Hyper-parameters} \quad For P-tuning, we prepend 128 soft prompt tokens to the input. The learnable prompts are randomly initialized and use the LSTM prompt encoder. For P-tuning-V2, each layer's soft prompt is directly randomly initialized without prompt encoder. The prompt length is also 128 for P-tuning-V2. Adapter is added in parallel to the feed-forward and self-attention module of the LLM, with a bottleneck dimension of 128. LoRA's rank is set to 24 and is added to the query, key, value, output weight matrix, and the two weight matrix in the feed-forward module. AdaLoRA's initial rank setting is the same with LoRA.

\textbf{Training settings} \quad Each PEFT method is trained with the same training hyper-parameters: warm up steps are 100, batch size is 64, and the learning rate is 3e-4, the optimizer is AdamW, the learning schedule is linear decay, gradient clipping is 1, weight decay is 1e-4, and the maximum epoch is 10.  The other settings follow the huggingface Transformers' default settings. We find that training a few steps after the dev loss hits the lowest point is helpful for generating better responses. Thus we pick the checkpoint that is 200 steps after the one with the lowest dev loss.

\subsection{Example dialogue for Table \ref{tab:error_mrg}}
\label{subsec:example_dialogue}

The complete patient-doctor conversation for Table \ref{tab:error_mrg} is as follows:
\begin{verbatim} 
患者：四个月宝宝血小板701有问题吗？宝宝最近有点咳嗽，去医院查血，白细胞十点几，血小板701，医生说怕是血栓。宝宝有点贫血，还有点缺钙，可能是因为奶水不是很多。昨天我感冒了，不知道有没有传染给宝宝。血小板高还可能是血液病，听一些网上的医生说这样的情况不用干预，会自己好的？？请专业医生解答一下，宝宝的问题大吗？应该怎么办呢？
医生：你好
患者：你好
患者：血小板高有问题吗？
医生：这个孩子的血小板偏高，应该与感染有一定的关系
医生：病毒感染或者说是炎症很容易导致血小板升高
医生：这个血小板数值达不到血小板增高症的诊断标准，一般来说超过八百，甚至说九百，才能诊断为血小板增高症
患者：有可能是血栓吗？
医生：血小板偏高会导致血液的粘滞度增加，发生血栓的概率比正常的孩子大的多，但是这种概率比较低
患者：问题大吗？
医生：我看你提供的病史，妈妈有感冒的情况，应该考虑被妈妈传染了
医生：血液病的可能性不大，应该考虑感染有一定的关系
医生：一般来说，这个血小板数值是不需要特殊处理的，随着孩子感染的控制，血小板的数值会逐渐的下降
患者：那怎么办？
医生：给孩子看感染治疗就可以了，因为这个孩子白细胞总数偏高啊，感染控制以后复查血常规
患者：问题不大吗
医生：是的
患者：嗯，谢谢！
医生：你好
患者：曲大夫你好
医生：三七可以吃个，具有活血化瘀的作用
患者：孩子今晚拉肚子了，看来炎症在肠胃。
医生：应该考虑胃肠道的
患者：可是才四个月大，能吃吗
医生：中药制剂的咳嗽药物可能会导致孩子出现拉肚子
医生：不建议口服的，这个孩子太小了
医生：如果就像口服的话，可以选择阿司匹林或者是潘生丁，这两个药物具有抑制血小板聚集的作用
患者：哦，孩子有点贫血
医生：血红蛋白是多少
患者：还没复查呢
医生：哦
患者：嗯，谢谢医生！
医生：这个孩子最主要的问题，是感染所导致的血小板升高，白细胞升高
\end{verbatim}

And its English translation is:
\begin{verbatim} 
Patient: Is there any problem with the platelet 701 of the four-month-old 
baby? The baby had a cough recently. I went to the hospital for a 
blood test. The white blood cells were over 10 and the platelets 
were 701. The doctor said it was probably a blood clot. The baby 
is a little anemic and a little calcium deficient, maybe because 
there is not a lot of milk. I caught a cold yesterday and I don’t 
know if it was passed on to my baby. High platelets may also be a 
blood disease. I heard some doctors on the Internet say that this 
kind of situation will heal on its own without intervention? ? 
Please ask a professional doctor to answer. Is the baby’s problem 
serious? How should I do it?
Doctor: Hello
Patient: Hello
Patient: Is there a problem with high platelets? 
Doctor: This child's platelets are high, which should be related 
to infection
Doctor: Viral infection or inflammation can easily lead to 
elevated platelets
Doctor: This platelet value does not meet the diagnostic criteria 
for thrombocytosis. , generally speaking, it can be diagnosed as 
thrombocytosis if it exceeds 800, or even 900.
Patient: Is it possible to have a blood clot? 
Doctor: High platelets will increase blood viscosity, and the 
probability of thrombosis is much higher than that of normal 
children, but this probability is relatively low
Patient: Is it a big problem? 
Doctor: From the medical history you provided, my mother has a 
cold, so I should consider that she was infected by her
Doctor: The possibility of blood disease is unlikely, and I should 
consider that infection is related to it.
Doctor: Generally speaking, this platelet value does not require 
special treatment. As the child's infection is controlled, the 
platelet value will gradually decrease
Patient: What should I do? 
Doctor: Just treat the child for infection, because the total number 
of white blood cells in this child is high. After the infection is 
controlled, the blood routine will be rechecked
Patient: Is it not a big problem?
Doctor: Yes
Patient: Well, thank you! 
Doctor: Hello
Patient: Hello, Dr. Qu
Doctor: You can eat Panax notoginseng, which has the effect of 
promoting blood circulation and removing blood stasis
Patient: My child has diarrhea tonight. It seems that the 
inflammation is in the intestines and stomach. 
Doctor: Gastrointestinal problems should be considered
Patient: But he is only four months old, can he take it?
Doctor: Traditional Chinese medicine preparations for cough medicine 
may cause diarrhea in children
Doctor: It is not recommended to take it orally because the child 
is too young
Doctor: If it is taken orally, you can choose aspirin or dipyridamole. 
These two drugs have inhibitory effects. The role of platelet 
aggregation
Patient: Oh, the child is a little anemic
Doctor: What is the hemoglobin
Patient: I haven’t reviewed it yet
Doctor: Oh
Patient: Well, thank you doctor! 
Doctor: The main problem with this child is the increase in platelets 
and white blood cells caused by infection.
\end{verbatim}

\bibliography{reference}
\bibliographystyle{iclr2024_conference}

\end{CJK*}